\documentclass[10pt,twocolumn,letterpaper]{article}

\usepackage{cvpr}
\usepackage{times}
\usepackage{epsfig}
\usepackage{graphicx}
\usepackage{amsmath}
\usepackage{amssymb}

\usepackage{subfig}
\usepackage{caption}
\usepackage{makecell}
\usepackage{booktabs}
\usepackage{subfig}

\usepackage{algorithm}
\usepackage[noend]{algpseudocode}

\usepackage{multicol}
\usepackage{multirow}

\usepackage[font=small,labelfont=bf,tableposition=top]{caption}
\usepackage{blindtext}
\usepackage{cuted}

\usepackage[numbers]{natbib}

\usepackage[pagebackref=true,breaklinks=true,letterpaper=true,colorlinks,bookmarks=false]{hyperref}

\cvprfinalcopy 

\def\cvprPaperID{3065} 

\ifcvprfinal\fi
\begin{document}

\title{SER-FIQ: Unsupervised Estimation of Face Image Quality Based on\\ Stochastic Embedding Robustness}

\newcommand\Mark[1]{\textsuperscript#1}

\author{Philipp Terh\"{o}rst\Mark{1}\Mark{2}, Jan Niklas Kolf\Mark{2}, Naser Damer\Mark{1}\Mark{2}, Florian Kirchbuchner\Mark{1}\Mark{2}, Arjan Kuijper\Mark{1}\Mark{2}\\
\Mark{1}Fraunhofer Institute for Computer Graphics Research IGD, Darmstadt, Germany\\
\Mark{2}Technical University of Darmstadt, Darmstadt, Germany\\
Email:{\{philipp.terhoerst, naser.damer, florian.kirchbuchner, arjan.kuijper\}@igd.fraunhofer.de}
}

\maketitle

\begin{strip}
\centering\noindent
\includegraphics[width=0.7\textwidth]{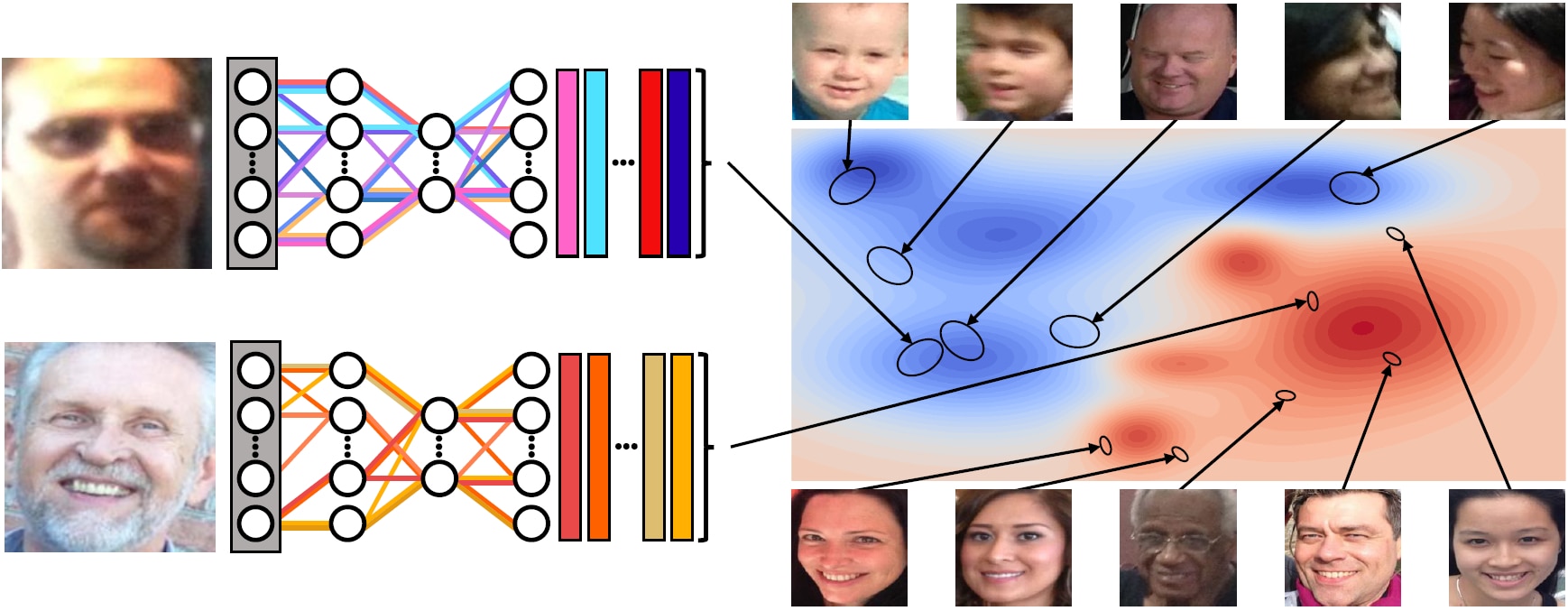}
\captionof{figure}{
Visualization of the proposed unsupervised face quality assessment concept. 
We propose using the robustness of an image representation as a quality clue.
Our approach defines this robustness based on the embedding variations of random subnetworks of a given face recognition model.
An image that produces small variations in the stochastic embeddings (bottom left), demonstrates   high robustness (red areas on the right) and thus, high image quality.
Contrary, an image that produces high variations in the stochastic embeddings (top left) coming from random subnetworks, indicates a low robustness (blue areas on the right).
Therefore, it is considered as low quality.
}
\label{fig:Visualization}
\end{strip}


\begin{abstract}
Face image quality is an important factor to enable high-performance face recognition systems.
Face quality assessment aims at estimating the suitability of a face image for recognition.
Previous work proposed supervised solutions that require artificially or human labelled quality values.
However, both labelling mechanisms are error-prone as they do not rely on a clear definition of quality and may not know the best characteristics for the utilized face recognition system.
Avoiding the use of inaccurate quality labels, we proposed a novel concept to measure face quality based on an arbitrary face recognition model.
By determining the embedding variations generated from random subnetworks of a face model, the robustness of a sample representation and thus, its quality is estimated.
The experiments are conducted in a cross-database evaluation setting on three publicly available databases. 
We compare our proposed solution on two face embeddings against six state-of-the-art approaches from academia and industry.
The results show that our unsupervised solution outperforms all other approaches in the majority of the investigated scenarios.
In contrast to previous works, the proposed solution shows a stable performance over all scenarios.
Utilizing the deployed face recognition model for our face quality assessment methodology avoids the training phase completely and further outperforms all baseline approaches by a large margin.
Our solution can be easily integrated into current face recognition systems and can be modified to other tasks beyond face recognition.

\end{abstract}

\section{INTRODUCTION}


Face images are one of the most utilized biometric modalities \cite{Wang2018} due to its high level of public acceptance and since it does not require an active user-participation \cite{Tripathi2017}. 
Under controlled conditions, current face recognition systems are able to achieve highly accurate performances \cite{FRVT2018}.
However, some of the most relevant face recognition systems work under unconstrained environments and thus, have to deal with large variabilities that leads to significant degradation of the recognition accuracies \cite{FRVT2018}.
These variabilities include image acquisition conditions (such as illumination, background, blurriness, and low resolution), factors of the face (such as pose, occlusions and expressions) \cite{ISO19794-5-2011, ICAO9303} and biases of the deployed face recognition system.
Since these variabilities lead to significantly degraded recognition performances, the ability to deal with these factors needs to be addressed \cite{DBLP:journals/corr/abs-1904-01740}.

The performance of biometric recognition is driven by the quality of its samples \cite{DBLP:journals/corr/Best-RowdenJ17}. 
Biometric sample quality is defined as the utility of a sample for the purpose of recognition \cite{DBLP:journals/corr/abs-1904-01740, 6712715, 10.1007/978-3-540-74549-5_26, DBLP:journals/corr/Best-RowdenJ17}.
The automatic prediction of face quality (prior to matching) is beneficial for many applications.
It leads to a more robust enrolment for face recognition systems.
In negative identification systems, it prevents an attacker from getting access to a system by providing a low quality face image.
Furthermore, it enables quality-based fusion approaches when multiple images \cite{DBLP:conf/icpr/DamerSN14} (e.g. from surveillance videos) or multiple biometric modalities are given.

Current solutions for face quality assessment require training data with quality labels coming from human perception or are derived from comparison scores.
Such a quality measure is generally poorly defined.
Humans may not know the best characteristics for the utilized face recognition system.
On the other hand, automatic labelling based on comparison scores represents the relative performance of two samples and thus, one low-quality sample might negatively affect the quality labels of the other one.

In this work, we propose a novel unsupervised face quality assessment concept by investigating the robustness of stochastic embeddings.
Our solution measures the quality of an image based on its robustness in the embedding space.
Using the variations of embeddings extracted from random subnetworks of the utilized face recognition model, the representation robustness of the sample and thus, its quality is determined.
Figure \ref{fig:Visualization} illustrates the working principle.

We evaluated the experiments on three publicly available databases in a cross-database evaluation setting. 
The comparison of our approach was done on two face recognition systems against six state-of-the-art solutions:
three no-reference image quality metrics, two recent face quality assessment algorithms from previous work, and one commercial off-the-shelf (COTS) face quality assessment product from industry.

The results show that the proposed solution is able to outperform all state-of-the-art solutions in most investigated scenarios.
While every baseline approach shows performance instabilities in at least two scenarios, our solution shows a consistently stable performance.
When using the deployed face recognition model for the proposed face quality assessment methodology, our approach outperforms all baseline by a large margin.
Contrarily to previous definitions of face quality assessment \cite{DBLP:journals/corr/Best-RowdenJ17,ISO19794-5-2011,ICAO9303,DBLP:journals/corr/abs-1904-01740} that states the face quality as a utility measure of a face image for an \textit{arbitrary} face recognition model, our results show that it is highly beneficial to estimate the sample quality with regard to a specific (the deployed) face recognition model.

%




\section{Related work}
\label{sec:RelatedWork}



%
%
%
%


Several standards have been proposed for insure face image quality by constraining the capture requirements, such as ISO/IEC 19794-5 \cite{ISO19794-5-2011} and ICAO 9303 \cite{ICAO9303}.
In these standards, quality is divided into \textit{image-based} qualities  (such as pose, expression, illumination, occlusion) and \textit{subject-based} quality measures (such as accessories).
These mentioned standards influenced many face quality assessment approaches that have been proposed in the recent years.
While the first solutions to face quality assessment focused on analytic image quality factors, current solutions make use of the advances in supervised learning.

Approaches based on analytic image quality factors define quality metrics for facial asymmetries \cite{10.1007/978-3-540-74549-5_26, 6197711}, propose vertical edge density as a quality metric to capture pose variations \cite{7935089}, or measured in terms of luminance distortion in comparison to a known reference image \cite{5424029}.
However, these approaches have to consider every possible factor manually, and since humans may not know the best characteristics for face recognition systems, more current research focus on learning-based approaches.

The transition to learning-based approaches include works that combine different analytical quality metrics with traditional machine learning approaches \cite{6712715, 6985846, 4341617, 6460821, 6996248}.

End-to-end learning approaches for face quality assessment were first presented in 2011.
Aggarwal et al. \cite{5981784} proposed an approach for predicting the face recognition performance using a multi-dimensional scaling approach to map space characterization features to genuine scores.
In \cite{5981881}, a patch-based probabilistic image quality approach was designed that works on 2D discrete cosine transform features and trains a Gaussian model on each patch.
In 2015, a rank-based learning approach was proposed by Chen et al. \cite{6877651}.
They define a linear quality assessment function with polynomial kernels and train weights based on a ranking loss.
In \cite{7351562}, face images assessment was performed based on objective and relative face image qualities.
While the objective quality metric refers to objective visual quality in terms of pose, alignment, blurriness, and brightness, the relative quality metric represents the degree of mismatch between training face images and a test face image.
Best-Rowden and Jain \cite{DBLP:journals/corr/Best-RowdenJ17} proposed an automatic face quality prediction approach in 2018.
They proposed two methods for quality assessment of face images based on (a) human assessments of face image quality and (b) quality values from similarity scores.
Their approach is based on support vector machines applied to deeply learned representations.
In 2019, Hernandez-Ortega et al. proposed FaceQnet \cite{DBLP:journals/corr/abs-1904-01740}.
This solution fine-tunes a face recognition neural network to predict face qualities in a regression task.
Beside image quality estimation for face recognition, quality estimation has been also developed to predict soft-biometric decision reliability based on the investigated image \cite{DBLP:conf/btas/Terhoerst19}.

All previous face image quality assessment solutions require training data with artificial or manually labelled quality values.
Human labelled data might transfer human bias into the quality predictions and does not take into account the potential biases of the biometric system.
Moreover, humans might not know the best quality factors for a specific face recognition system.
Artificially labelled quality values are created by investigating the relative performance of a face recognition system (represented by comparison scores).
Consequently, the score might be heavily biased by low-quality samples.

The solution presented in this paper is based on our hypothesis that representation robustness is better suited as a quality metric, since it provides a measure for the quality of a single sample independently of others and avoids the use of misleading quality labels for training. 
This metric can intrinsically capture image acquisition conditions and factors of the face that are relevant for the used face recognition system.
Furthermore, it is not affected by human bias, but takes into account the bias and the decision patterns of the used face embeddings.

\section{Our approach}







\begin{figure}[h]
\centering
\includegraphics[width=0.4\textwidth]{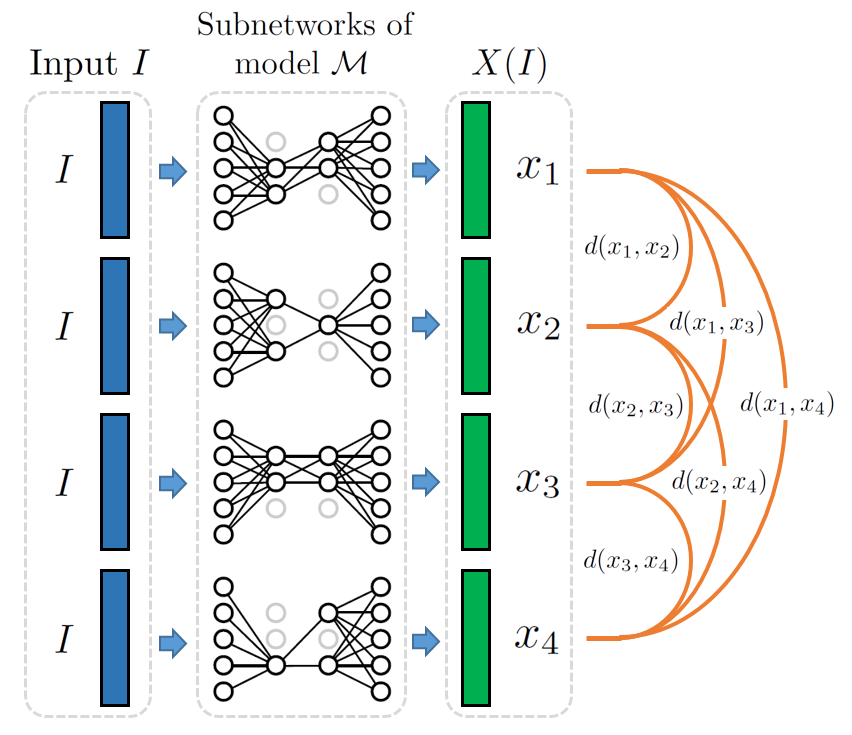}
\caption{Illustration of the proposed methodology: an input $I$ is forwarded to different random subnetworks of the used face recognition model $\mathcal{M}$. Each subnetwork produces a different stochastic embedding $x_s$. The variations between these embeddings are calculated using pairwise-distances and define the quality of $I$. }
\label{fig:Illustration}
\end{figure}

Face quality assessment aims at estimating the suitability of a face image for face recognition.
The quality of a face image should indicate its expected recognition performance.
In this work, we based our face image quality definition on the relative robustness of deeply learned embeddings of that image.
Calculating the variations of embeddings coming from random subnetworks of a face recognition model, our solution defines the magnitude of these variations as a robustness measure, and thus, image quality.
An illustration of this methodology is shown in Figure \ref{fig:Illustration}. 

%
%
%
%
%
%
%
%
%
%
%
%


\subsection{Sample-quality estimation}

More formally, our proposed solution predicts the face quality $Q(I)$ of a given face image $I$ using a face recognition model $\mathcal{M}$.
The face recognition model have to be trained with dropout and aims at extracting embeddings that are well identity-separated.
To make a robustness-based quality estimation of $I$, $m=100$ stochastic embeddings are generated from the model $\mathcal{M}$ using stochastic forward passes with different dropout patterns.
The choice for $m$ is defined by the trade-off between time complexity and stability of the quality measure as described in Section \ref{sec:Properties}.
Each stochastic forward pass applies a different dropout pattern (during prediction) producing a different subnetwork of $\mathcal{M}$. 
Each of these subnetworks generates different stochastic face embeddings $x_s$.
These stochastic embeddings are collected in a set $X(I)=\{x_s\}_{s \in \left\lbrace 1,2,\dots,m\right\rbrace}$.
We define the face quality
\begin{align}
q(X(I)) = 2 \, \sigma \Big( - \dfrac{2}{m^2} \sum_{i<j} d(x_i, x_j) \Big),
\end{align}
of image $I$ as the sigmoid of the negative mean euclidean distance $d(x_i, x_j)$ between all stochastic embeddings pairs $(x_i, x_j)\in X \times X$.
The sigmoid function $\sigma(\cdot)$ ensures that  $q\in [0,1]$. 
Since Gal et al. \cite{Gal2015DropoutAA} proofed that applying dropout repetitively on a network approximates the uncertainty of a Gaussian process \cite{Rasmussen06gaussianprocesses}, the euclidean distance is a suitable choice for $d(x_i, x_j)$. 
A greater variation in the stochastic embedding set $X$ indicate a low robustness of the representation and thus, a lower sample quality $q$.
Lower variations in $X$ indicate high robustness in the embedding space and is considered as a high sample quality $q$.
The quality prediction strategy is summarized in Algorithm \ref{algor:SER}. 

\algnewcommand\algorithmicinput{\textbf{Input:}}
\algnewcommand\INPUT{\item[\algorithmicinput]}

\algnewcommand\algorithmicoutput{\textbf{Output:}}
\algnewcommand\OUTPUT{\item[\algorithmicoutput]}

\begin{algorithm}[h]
\caption{Stochastic Embedding Robustness (SER)}\label{algor:SER}
\begin{algorithmic}[1]
\INPUT preprocessed input image $I$, NN-model $\mathcal{M}$
\OUTPUT quality value $Q$ for input image $I$ 
\Procedure{SER}{$I$, $\mathcal{M}$, $m=100$}
\State $X \gets \text{empty list}$
\For{$i \gets 1, \dots, m$}
\State $x_i \gets \mathcal{M}.pred(I, dropout=True)$
\State $X = X.add(x_i)$
\EndFor
\State $Q \gets q(X)$
\State \Return $Q$
\EndProcedure
\end{algorithmic}
\vspace{-1mm}
\end{algorithm}

\subsection{Properties}
\label{sec:Properties}

The aim of SER-FIQ is to estimate the face image quality from the perspective of utilisation in recognition tasks, which might be different than estimating the notion of image quality.
An image that produces relatively stable identity-related embeddings despite various variations (here caused by dropout) is an image with high utilisation in a recognition task, given that the recognition network training aims at being robust against intra-identity variations.

Face recognition algorithms are trained with the aim of learning robust representations to increase inter-identity separability and decrease intra-identity separability.
Assuming that a face recognition network is trained with dropout and the quality of a sample correlates with its embedding robustness, different subnetworks can be created from the basic model so that they possess different dropout patterns. 
The agreement between the subnetworks can be used to estimate the embedding robustness, and thus the quality.
If the $m$ subnetworks produce similar outputs (high agreement), the variations over these random subnetworks (the stochastic embedding set $X$) are low. 
Consequently, the robustness of this embedding, and thus the quality of the sample, is high.
Conversely, if the $m$ subnetworks produce dissimilar representations (low agreement), the variations over the random subnetworks are high.
Therefore, the robustness in the embedding space is low and the quality of the sample can be considered low as well.

Our approach has only one parameter $m$, the number of stochastic forward passes.
This parameter can be interpreted as the number of steps in a Monte-Carlo simulation and controls the stability of the quality predictions.
A higher $m$ leads to more stable quality estimates.
Since the computational time $t = \mathcal{O}(m^2)$ of our method grows quadratically with $m$,
it should not be chosen too high.
However, our method can compensate for this issue and can easily run in real-time, since it is highly parallelizable and the computational effort can be greatly reduced by repeating the stochastic forward passes only through the last layer(s) of the network.

In contrast to previous work, our solution does not require quality labels for training.
Furthermore, if the deployed face recognition system was trained with dropout, the same network can be used for determining the embedding robustness and therefore, the sample quality.
By doing so the training phase can be completely avoided and the quality predictions further captures the decision patterns and bias of the utilized face recognition model. 
Therefore, we highly recommend utilizing the deployed face recognition model for the quality assessment task.




\section{Experimental setup} 
\label{sec:ExperimentalSetup}

\paragraph{Databases}
The face quality assessment experiments were conducted on three publicly available databases chosen to have variation in quality and to prove the generalization of our approach on multiple databases.
The ColorFeret database \cite{ColorFERET} consists of 14,126 high-resolution face images from 1,199 different individuals.
The data possess a variety of face poses and facial expressions under well-controlled conditions.
The Adience dataset \cite{Eidinger:2014:AGE:2771306.2772049} consists of 26,580 images from over 2,284 different subjects under unconstrained imaging conditions.
Labeled Faces in the Wild (LFW) \cite{LFWTech} contains 13,233 face images from 5749 identities.
For both datasets, large variations in illumination, location, focus, blurriness, pose, and occlusion are included.

\paragraph{Evaluation metrics}

To evaluate the face quality assessment performance, we follow the methodology by Grother et al. \cite{DBLP:journals/pami/GrotherT07} using error versus reject curves.
These curves show a verification error-rate over the fraction of unconsidered face images.
Based on the predicted quality values, these unconsidered images are these with the lowest predicted quality and the error rate is calculated on the remaining images.
Error versus reject curves indicates good quality estimation when the verification error decreases consistently when increasing the ratio of unconsidered images.
In contrast to error versus quality-threshold curves, this process allows to fairly compare different algorithms for face quality assessment, since it is independent of the range of quality predictions.
The cruve was adapted in the approved ISO working item \cite{ISO} and used in the literature \cite{DBLP:journals/corr/Best-RowdenJ17, DBLP:reference/bio/TabassiG15, VendorTest}.

The face verification error rates within the error versus reject curves are reported in terms of false non-match rate (FNMR) at fixed false match rate (FMR) and as equal error rate (EER).
The EER equals the FMR at the threshold where FMR = 1$-$FNMR and is well known as a single-value indicator of the verification performance.
These error rates are specified for biometric verification evaluation in the international standard \cite{ISO_Metrik}.
In our experiment, we report the face verification performance on three operating points to cover a wider range of potential applications.
The face recognition performance is reported in terms of EER and FNMR at a FMR threshold of 0.01.
The FNMR is also reported at 0.001 FMR threshold as recommended by the best practice guidelines for automated border control of Frontex \cite{FrontexBestPractice}.

\paragraph{Face recognition networks}
To get face embedding from a given face image, the image is aligned, scaled, and cropped. 
The preprocessed image is passed to a face recognition models to extract the embeddings.
In this work, we use two face recognition models, FaceNet \cite{DBLP:journals/corr/SchroffKP15} and ArcFace \cite{Deng_2019_CVPR}.
For FaceNet, the image is aligned, scaled, and cropped as described in \cite{Kazemi2014OneMF}.
To extract the embeddings, a pretrained model\footnote{\url{https://github.com/davidsandberg/facenet}} was used.
For ArcFace, the image preprocessing was done as described in \cite{DBLP:journals/corr/abs-1812-01936} and a pretrained model\footnote{\url{https://github.com/deepinsight/insightface}} provided by the authors of ArcFace is used.
Both models were trained on the MS1M database \cite{DBLP:journals/corr/GuoZHHG16}.
The output size is 128 for FaceNet and 512 for ArcFace.
The identity verification is performed by comparing two embeddings using cosine-similarity.

\paragraph{On-top model preparation}
To apply our quality assessment methodology, a recognition model that was trained with dropout \cite{Srivastava:2014:DSW:2627435.2670313} is needed.
Otherwise, a model containing dropout need to added on the top of the existing model.
The direct way to apply our approach is to take a pretrained recognition model and repeat the stochastic forward passes only in the last layer(s) during prediction.
This is even expected to reach a better performance than training a custom network, because the verification decision, as well as the quality estimation decision, is done in a shared embedding space.

To demonstrate that our solution can be applied to any arbitrary face recognition system, in our experiments we show both approaches:
(a) training a small custom network on top of the deployed face recognition system, which we will refer to as \textit{SER-FIQ (on-top model)}, and (b) using the deployed model for the quality assessment, which we will refer to as \textit{SER-FIQ (same model)}.

The structure of \textit{SER-FIQ (on-top model)} was optimized such that its produced embeddings achieve a similar EER on ColorFeret as that of the FaceNet embeddings.
It consist of five layers with $n_{emb}/128/512/n_{emb}/n_{ids}$ dimensions.
The two intermediate layers have 128 and 512 dimensions.
The last layer has the dimension equal to the number of training identities $n_{ids}$ and is only needed during training.
All layers contain dropout \cite{Srivastava:2014:DSW:2627435.2670313} with the recommended dropout probability $p_d=0.5$ and a tanh activation.
The training of the small custom network is done using the AdaDelta optimizer \cite{DBLP:journals/corr/abs-1212-5701} with a batchsize of 1024 over 100 epochs.
Since the size of the in- and output layers (blue and green) of the networks differs dependent on the used face embeddings, a learning rate of $\alpha_{FN}=10^{-1}$ was chosen for FaceNet and $\alpha_{AF}=10^{-4}$ for the higher dimensional ArcFace embeddings.
As the loss function, we used a simple binary cross-entropy loss on the classification of the training identities.

\paragraph{Investigations}
To investigate the generalization of face quality assessment performance, we conduct the experiments in a cross-database setting.
The training is done on ColorFeret to make the models learn variations in a controlled environment.
The testing is done on two unconstrained datasets, Adience and LFW.
The embeddings used for the experiments are from the widely used FaceNet (2015) and recently published ArcFace (2019) models.

To put the experiments in a meaningful setting, we evaluated our approach in comparison to six baseline solutions.
Three of these baselines are well-known no-reference image quality metrics from the computer vision community: Brisque \cite{6272356}, Niqe \cite{6353522}, Piqe \cite{7084843}.
The other three baselines are state-of-the-art face quality assessment approaches from academia and industry.
COTS \cite{COTS} is an off the shelf industry product from Neurotechnology.
We further compare our method with the two recent approaches from academia:
the face quality assessment approach presented by Best-Rowden and Jain \cite{DBLP:journals/corr/Best-RowdenJ17} (2018) and FaceQnet \cite{DBLP:journals/corr/abs-1904-01740} (2019).
Training the solution presented by Best-Rowden was done on ColorFeret following the procedure described in \cite{DBLP:journals/corr/Best-RowdenJ17}.
The generated labels come from cosine similarity scores using the same embeddings as in the evaluation scenario.
For all other baselines, pretrained models are utilized.

Our proposed methodology is presented in two settings,
the \textit{SER-FIQ (on-top model)} and the \textit{SER-FIQ (same model)}.
\textit{SER-FIQ (on-top model)} demonstrates that our unsupervised method can be applied to any face recognition system.
\textit{SER-FIQ (same model)} make use of the deployed face recognition model for quality assessment, to show the effect of capture its decision patterns for face quality assessment.
In the latter case, we apply the stochastic forward passes only between the last two layers of the deployed face recognition network. 


\begin{figure}[th]
\centering
  \subfloat[COTS \label{fig:quality_COTS}]{%
       \includegraphics[width=0.12\textwidth]{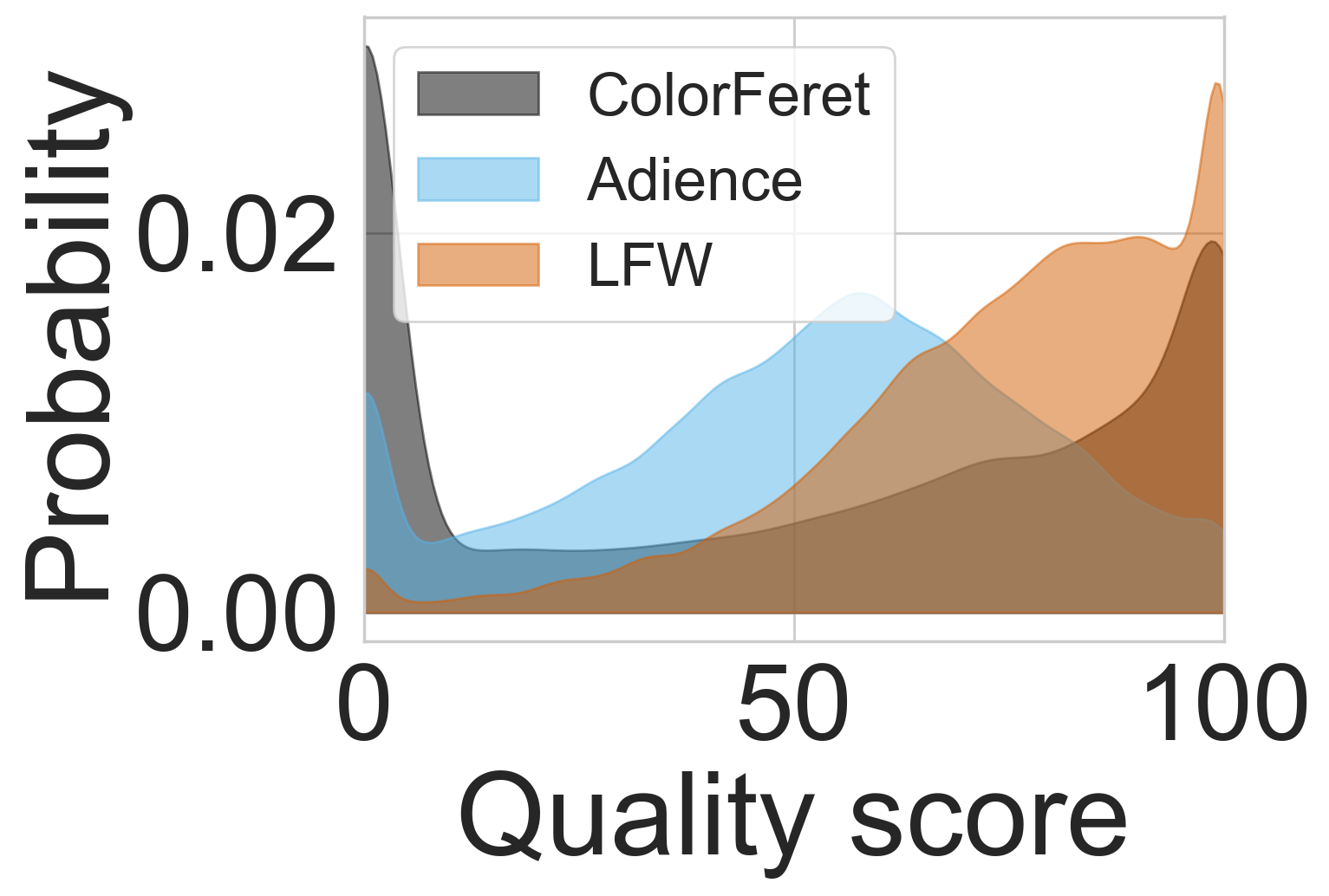}} 
  \subfloat[FaceQnet \label{fig:quality_faceQnet}]{%
       \includegraphics[width=0.12\textwidth]{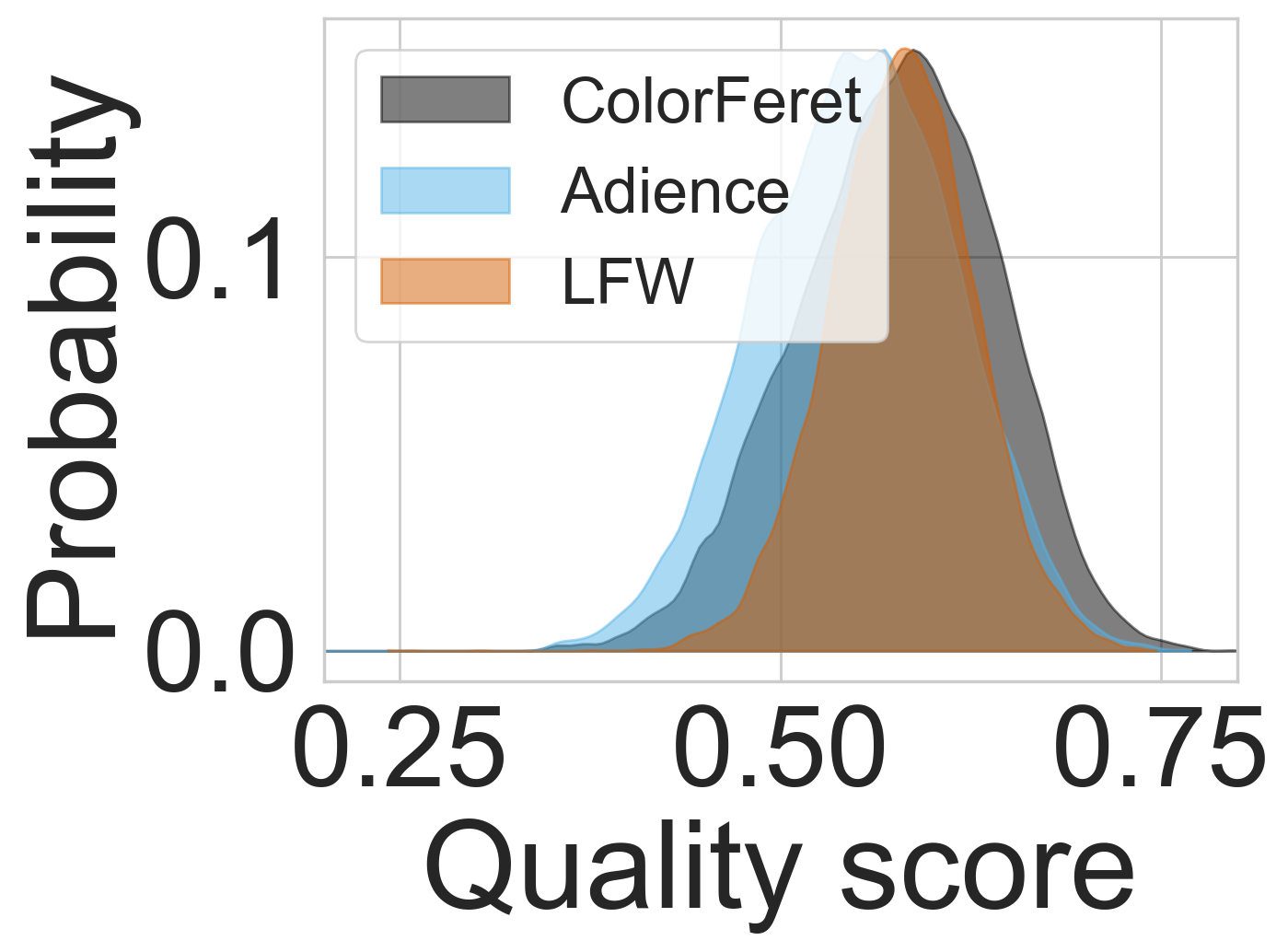}}
   \subfloat[SER-FIQ \newline \hspace{3mm} (on FaceNet) \label{fig:quality_SER-FIQ_FaceNet}]{%
       \includegraphics[width=0.12\textwidth]{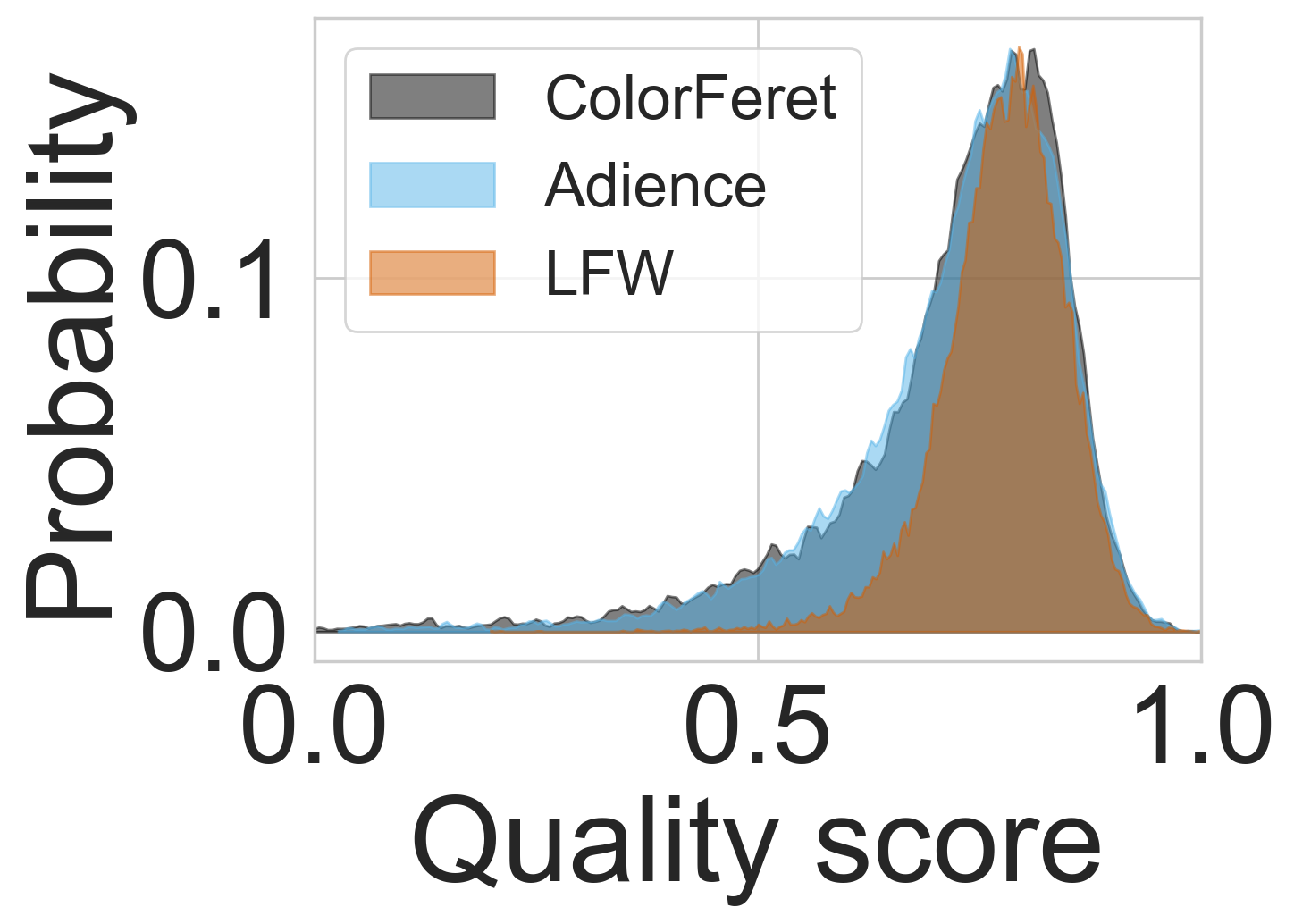}}
   \subfloat[SER-FIQ \newline (on ArcFace) \label{fig:quality_SER-FIQ_ArcFace}]{%
       \includegraphics[width=0.12\textwidth]{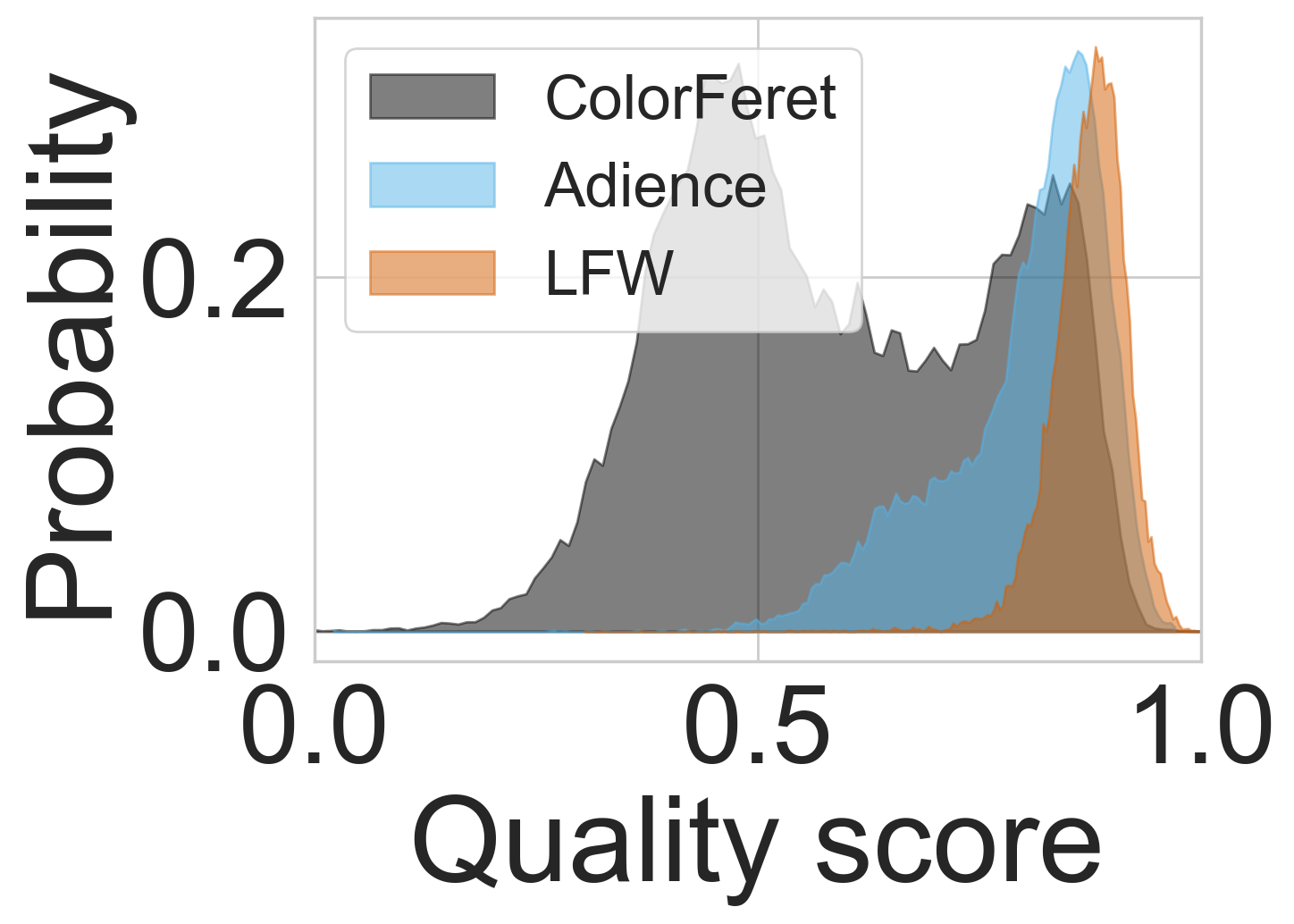}}
\caption{Face quality distributions of the used databases: Adience, LFW, and ColorFeret. The quality predictions were done using the pretrained models FaceQnet \cite{DBLP:journals/corr/abs-1904-01740}, COTS \cite{COTS}, and the proposed SER-FIQ (same model) based on FaceNet and ArcFace.} \vspace{-2mm}
\label{fig:qualityDistributions}
\end{figure}

\begin{figure*}[h]
\centering
  \subfloat[Adience - FaceNet \label{fig:001FMR_adience_facenet}]{%
       \includegraphics[width=0.43\textwidth]{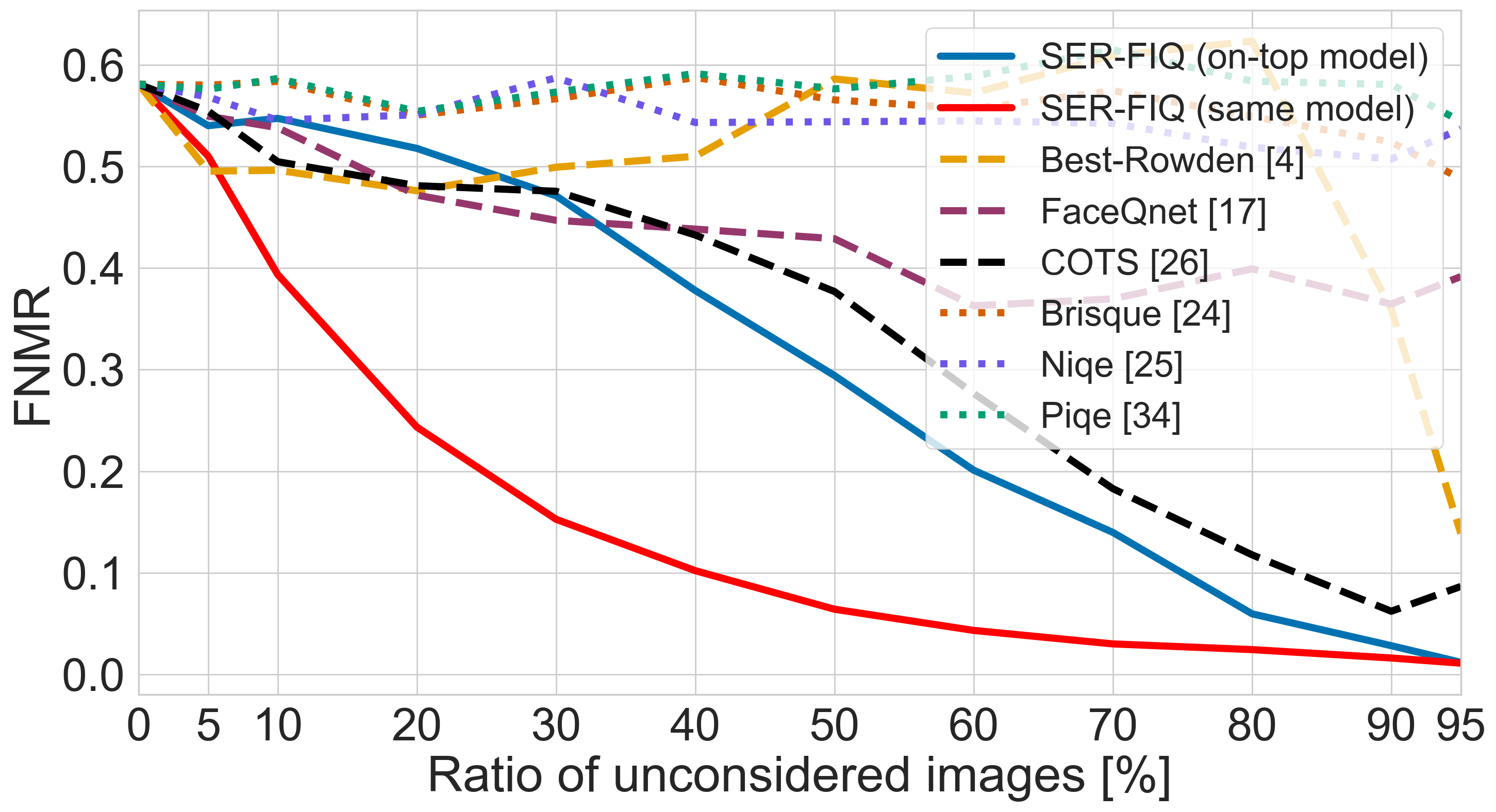}} \hspace{1mm}
  \subfloat[Adience - ArcFace \label{fig:001FMR_adience_arcface}]{%
       \includegraphics[width=0.43\textwidth]{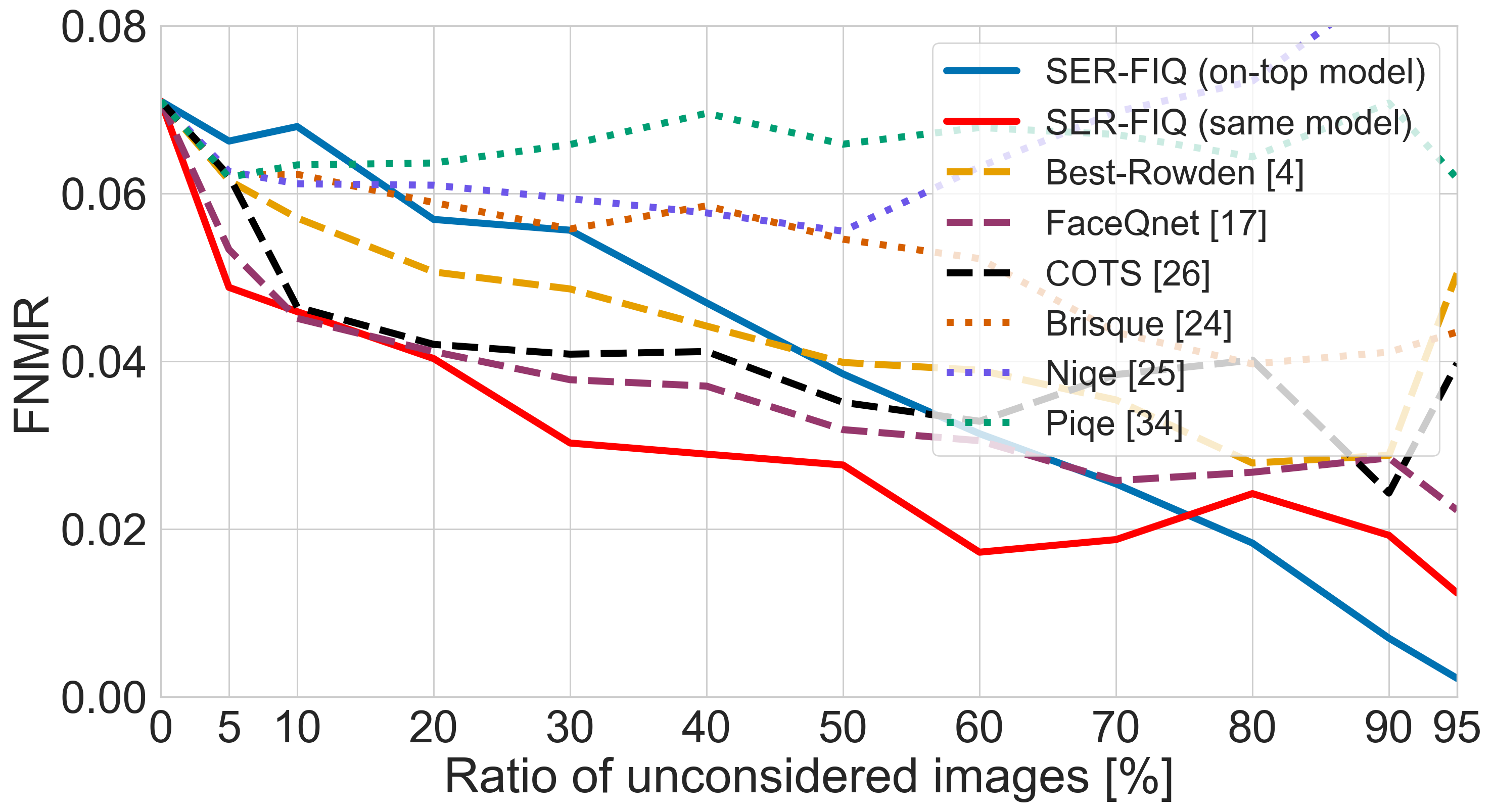}} \hspace{1mm}
  \subfloat[LFW - FaceNet \label{fig:001FMR_lfw_facenet}]{%
       \includegraphics[width=0.43\textwidth]{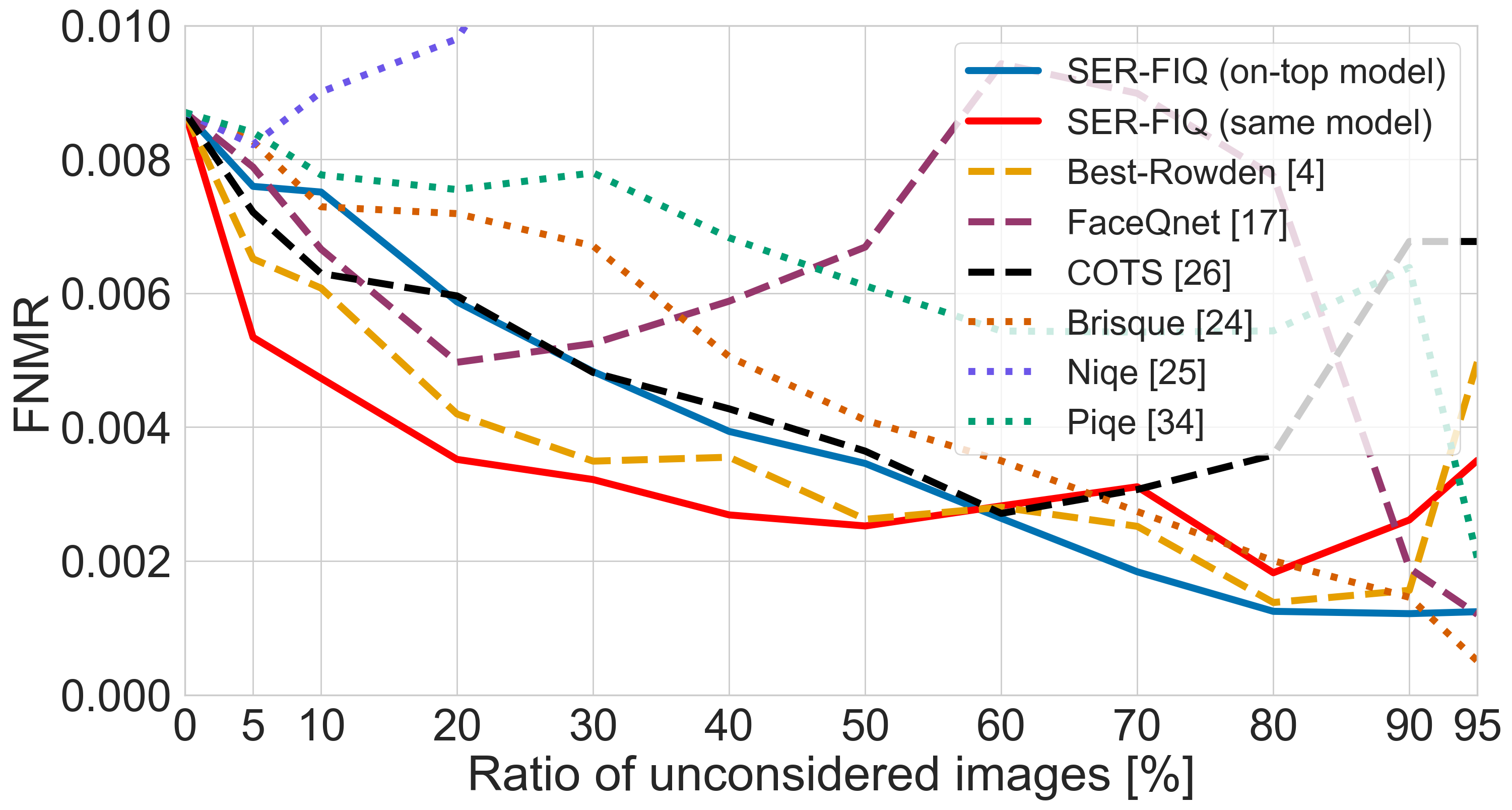}} \hspace{1mm}
  \subfloat[LFW - ArcFace \label{fig:001FMR_lfw_arcface}]{%
       \includegraphics[width=0.43\textwidth]{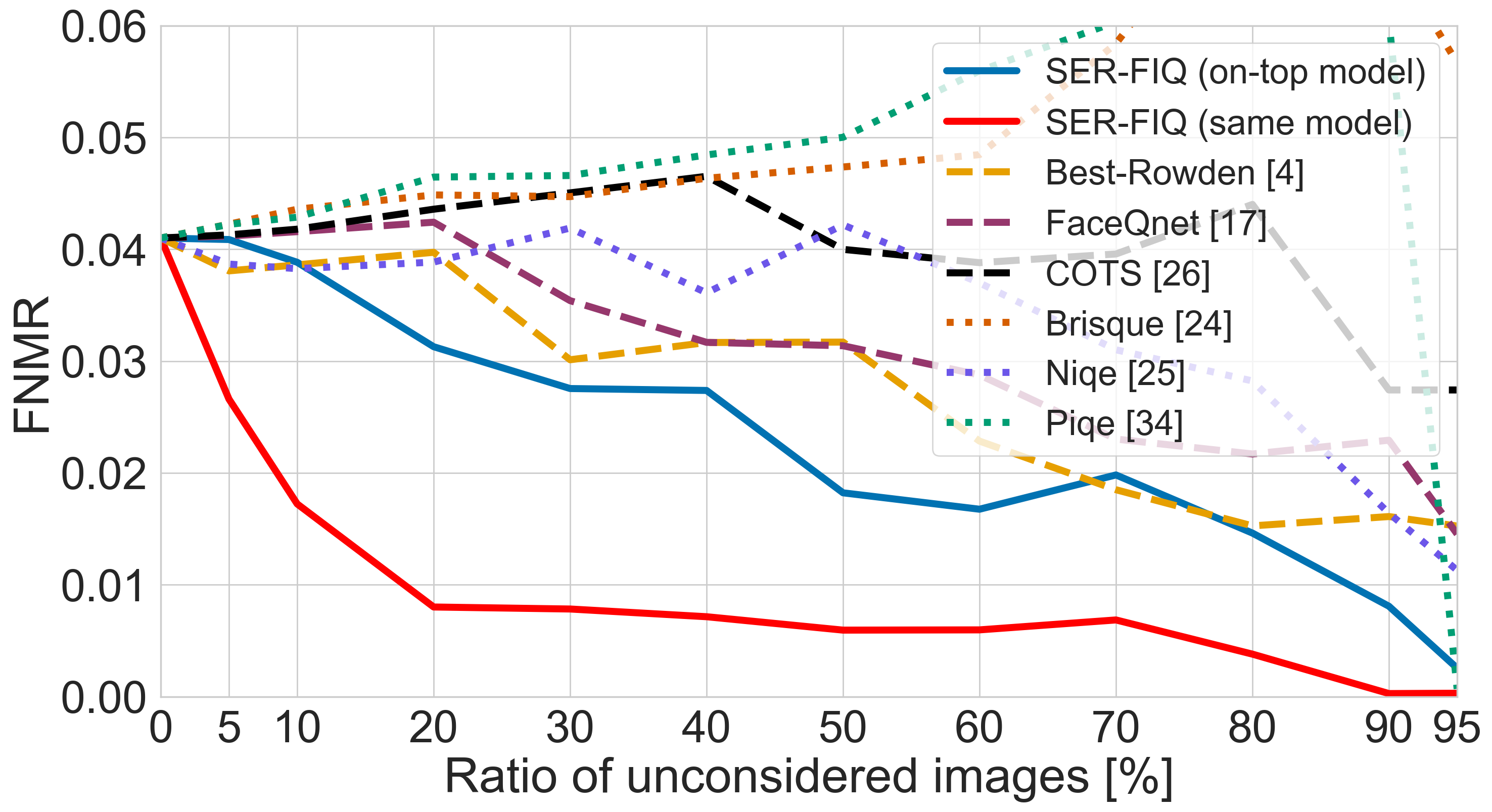}} \hspace{1mm}
\caption{Face verification performance for the predicted face quality values. The curves show the effectiveness of rejecting low-quality face images in terms of FNMR at a threshold of 0.001 FMR. Figure \ref{fig:001FMR_adience_facenet} and \ref{fig:001FMR_adience_arcface} show the results for FaceNet and ArcFace embeddings on Adience. Figure \ref{fig:001FMR_lfw_facenet} and \ref{fig:001FMR_lfw_arcface} show the same on LFW.} \vspace{-2mm}
\label{fig:001FMR}
\end{figure*}

\paragraph{Database face quality rating}
To justify the choices of the used databases, Figure \ref{fig:qualityDistributions} shows the face quality distributions of the databases using quality estimates from four pretrained face quality assessment models.
ColorFeret was captured under well-controlled conditions and generally shows very high qualities.
However, it contains non-frontal head poses and for COTS and SER-FIQ (on FaceNet) (Figure \ref{fig:quality_COTS}) this is considered as low image quality.
Because of these controlled variations, we choose ColorFeret as the training database.
Adience and LFW are unconstrained databases and for all quality measures, most face images are far away from perfect quality conditions.
For this reason, we choose these databases for testing.

%
%


\vspace{-0mm}

\section{Results}

\begin{figure}[h]
\centering\noindent
\includegraphics[width=0.090\textwidth]{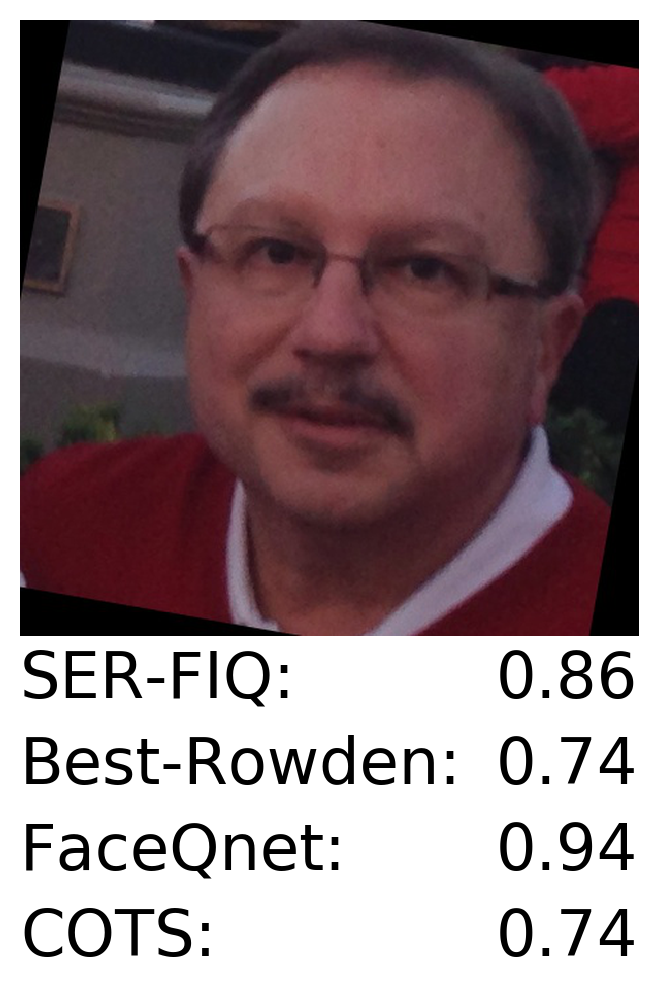}
\includegraphics[width=0.090\textwidth]{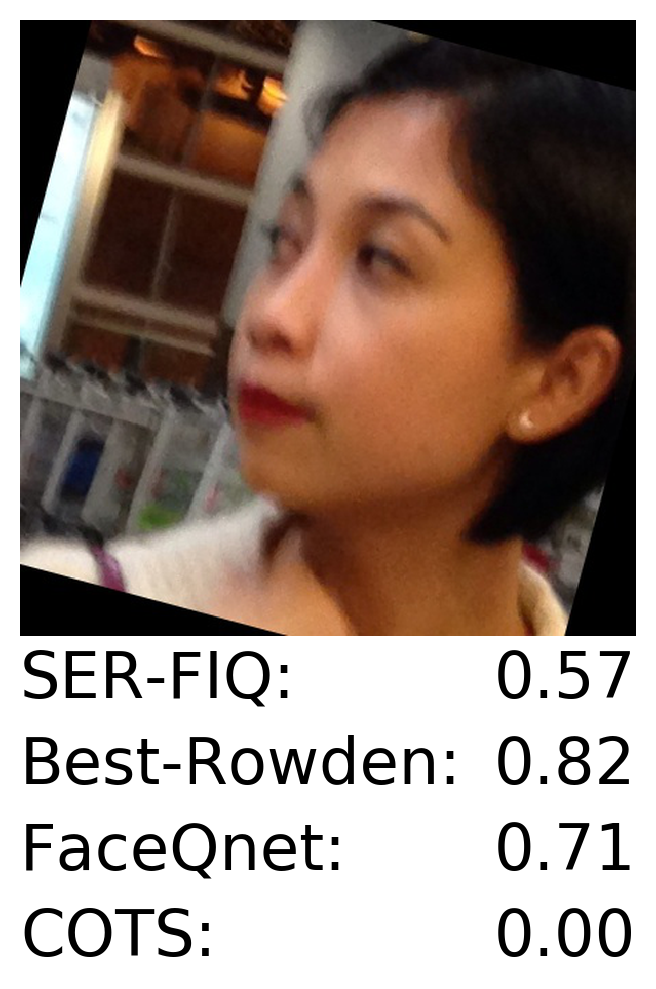}
\includegraphics[width=0.090\textwidth]{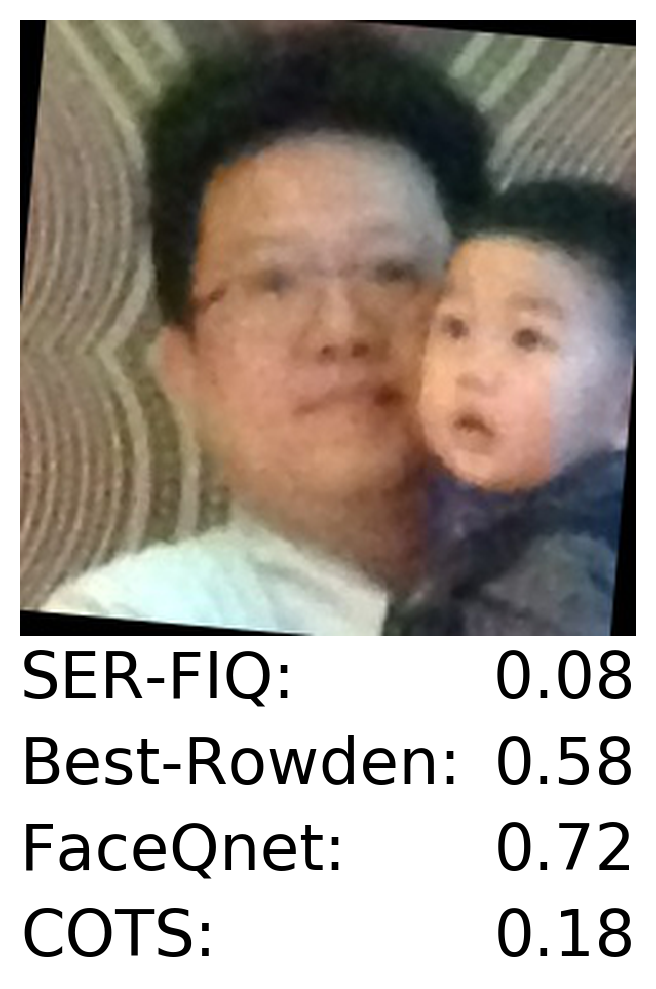}
\includegraphics[width=0.090\textwidth]{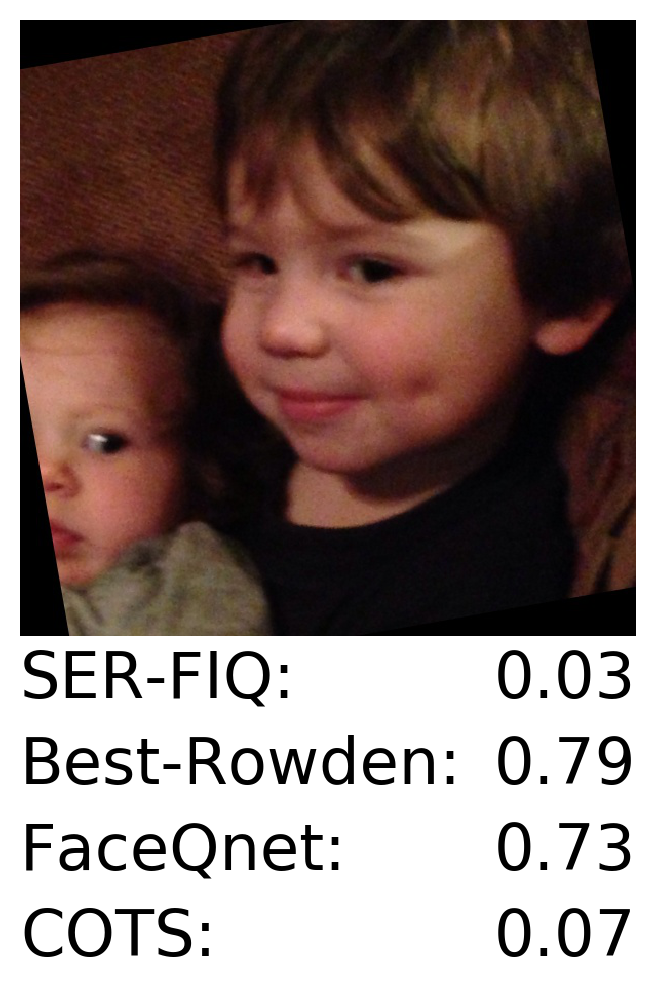}
\includegraphics[width=0.090\textwidth]{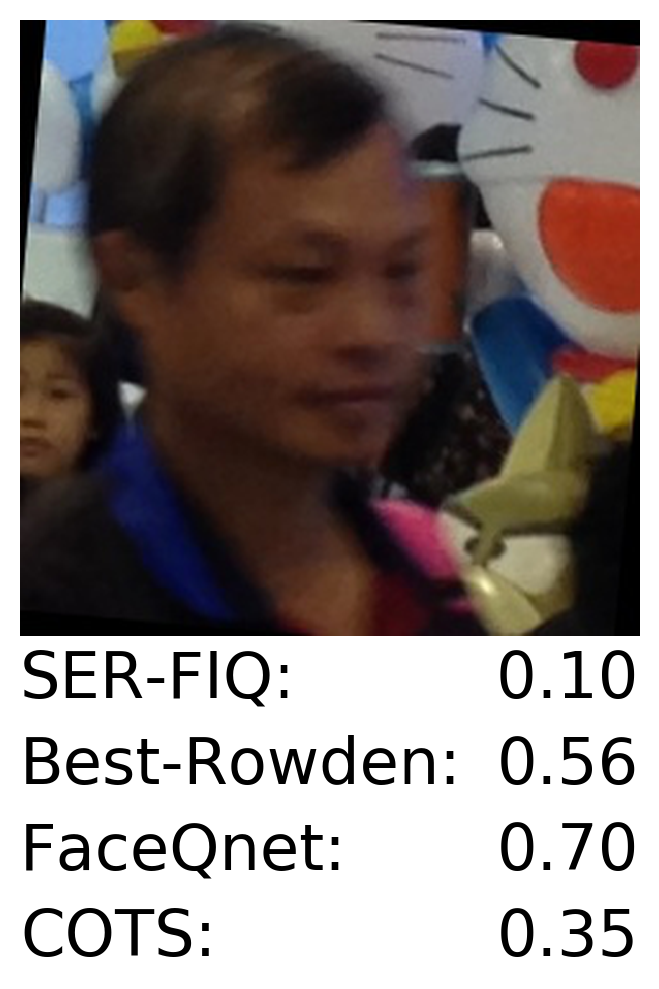}
\caption{Sample face images from Adience with the corresponding quality predictions from four face quality assessment methods. SER-FIQ refers to our same model approach based on ArcFace.}
\label{fig:SampleImages}
\end{figure}

\begin{figure*}[h]
\centering
  \subfloat[Adience - FaceNet \label{fig:01FMR_adience_facenet}]{%
       \includegraphics[width=0.43\textwidth]{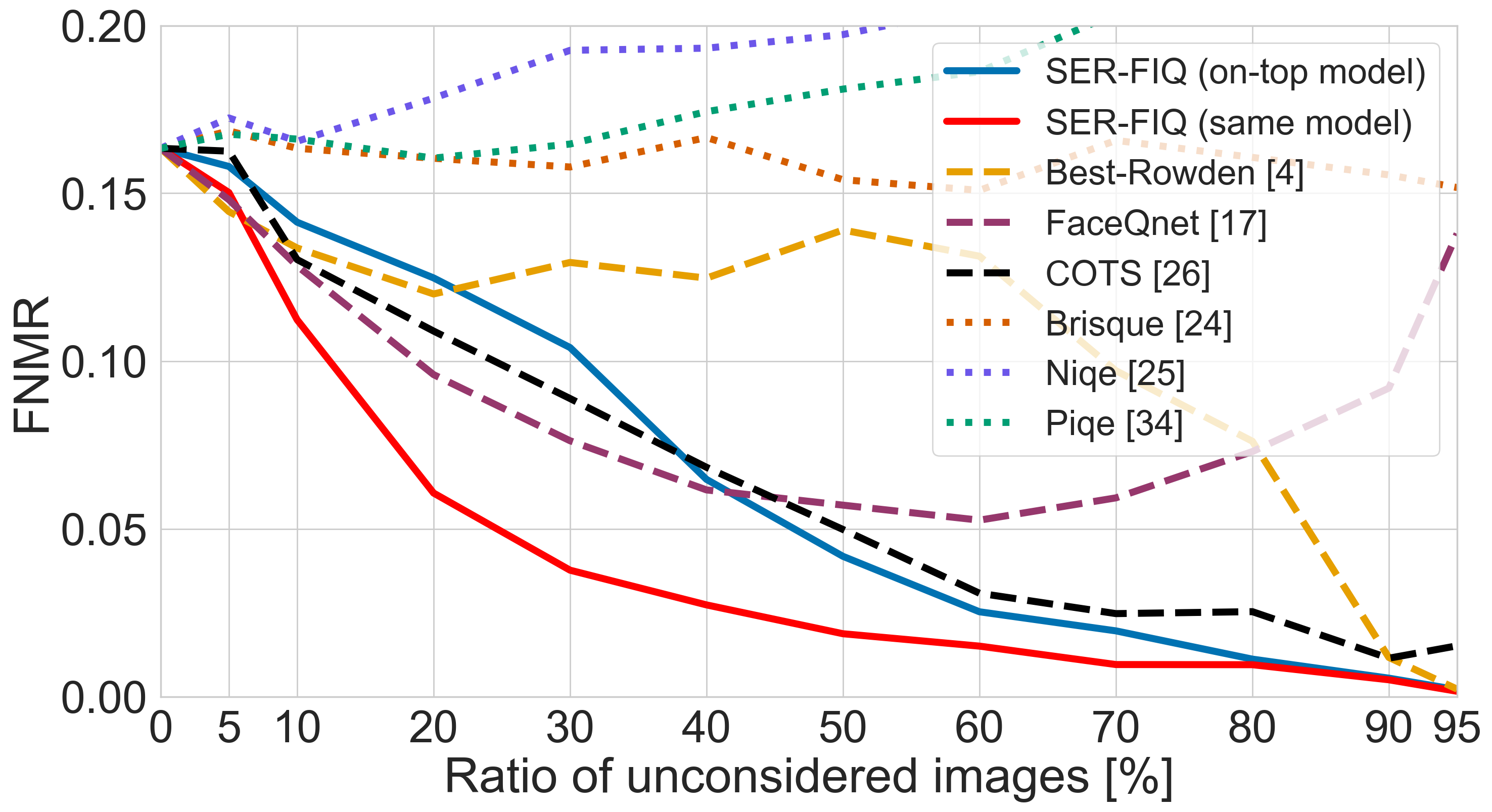}} \hspace{1mm}
  \subfloat[Adience - ArcFace \label{fig:01FMR_adience_arcface}]{%
       \includegraphics[width=0.43\textwidth]{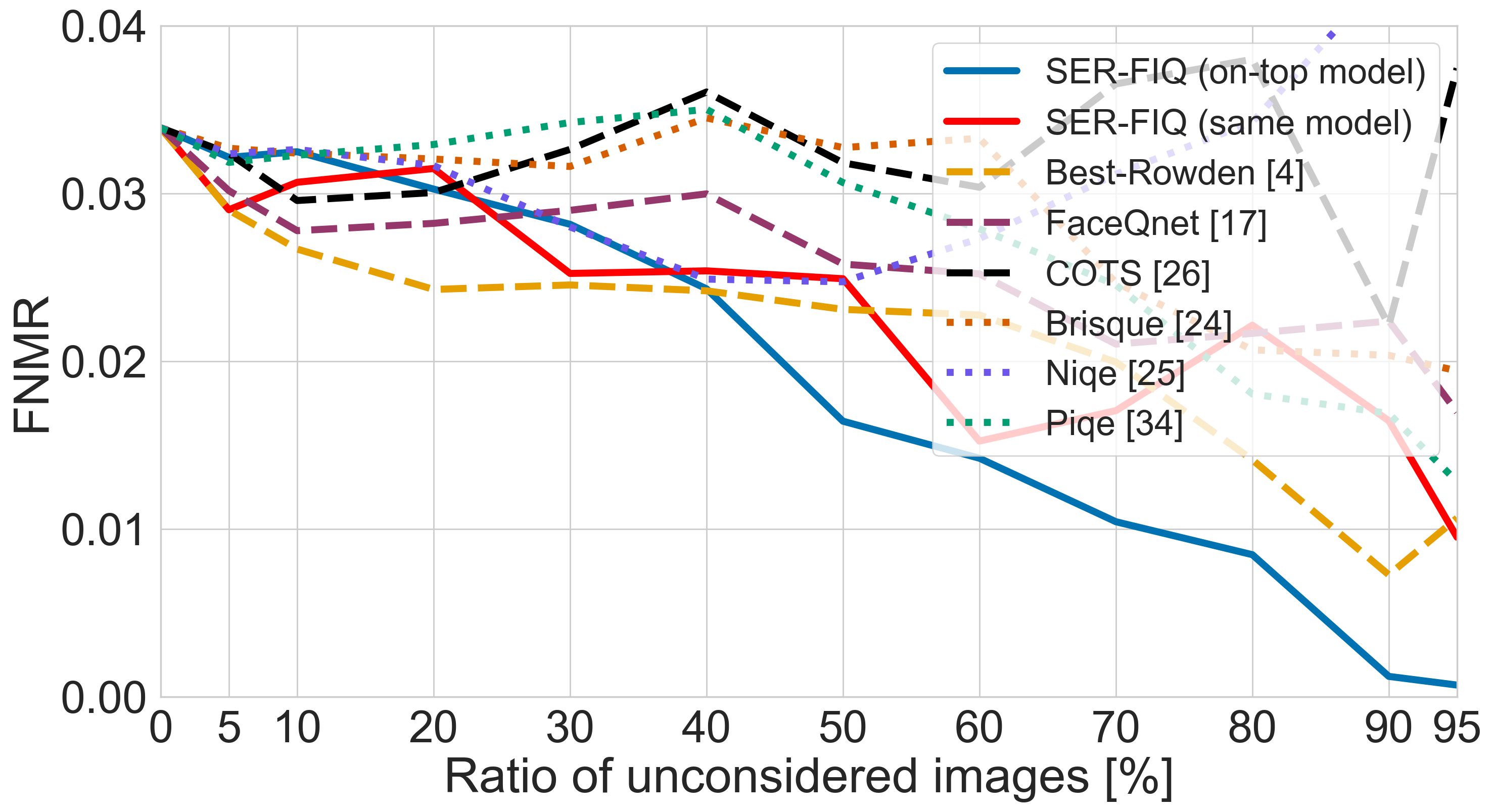}} \hspace{1mm}
  \subfloat[LFW - FaceNet \label{fig:01FMR_lfw_facenet}]{%
       \includegraphics[width=0.43\textwidth]{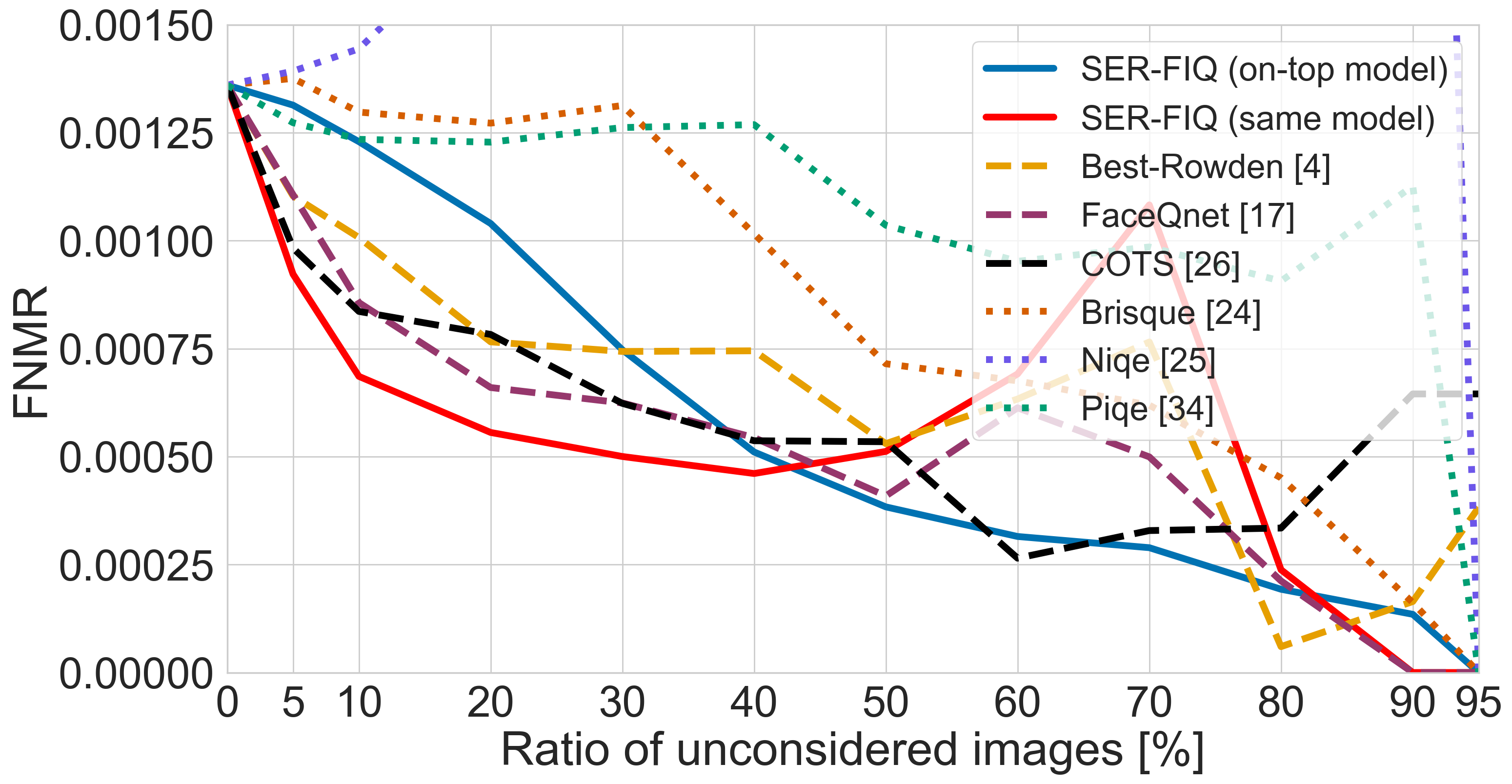}} \hspace{1mm}
  \subfloat[LFW - ArcFace \label{fig:01FMR_lfw_arcface}]{%
       \includegraphics[width=0.43\textwidth]{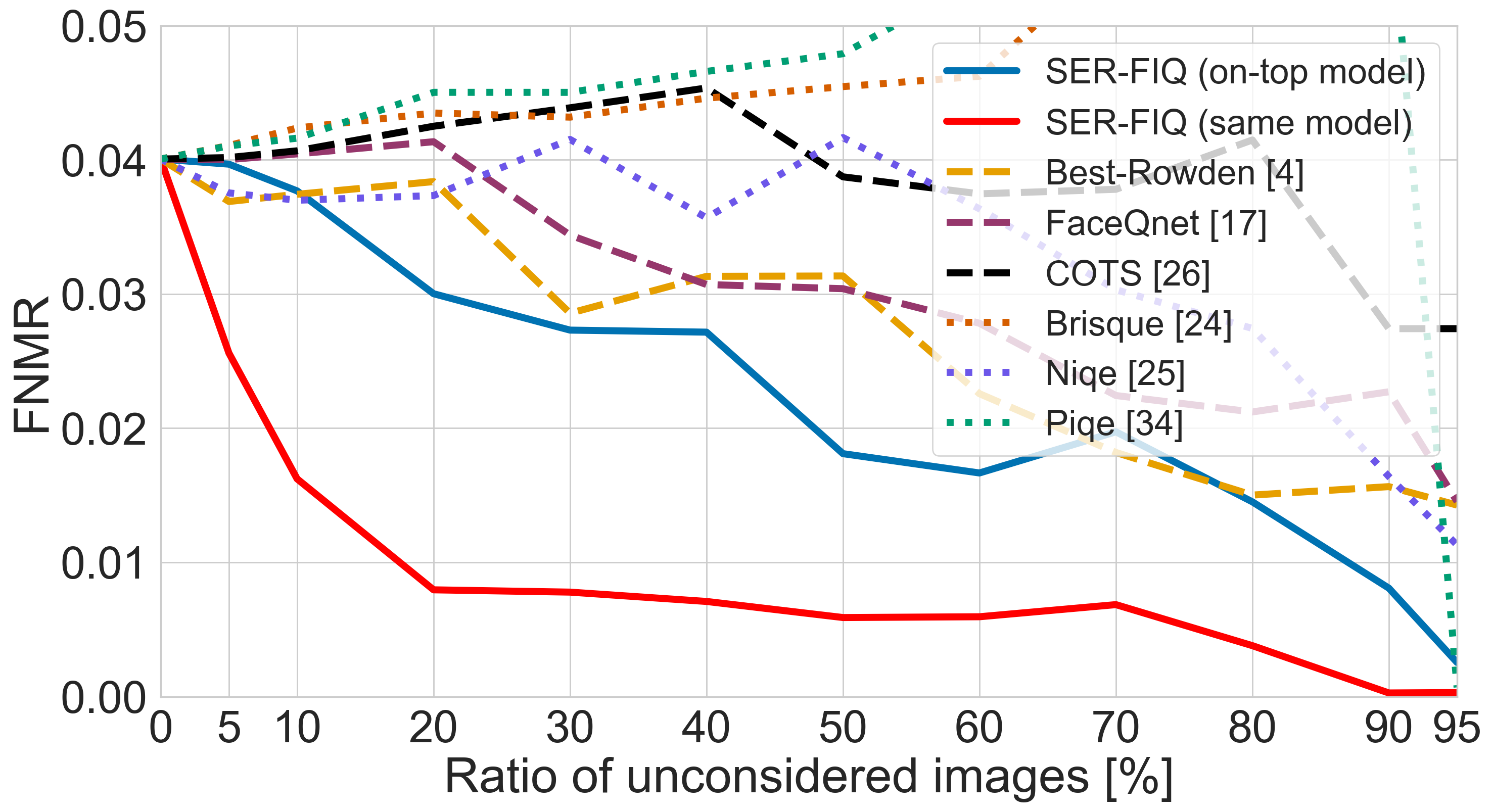}} \hspace{1mm}
\caption{Face verification performance for the predicted face quality values. The curves show the effectiveness of rejecting low-quality face images in terms of FNMR at a threshold of 0.01 FMR. Figure \ref{fig:01FMR_adience_facenet} and \ref{fig:01FMR_adience_arcface} show the results for FaceNet and ArcFace embeddings on Adience. Figure \ref{fig:01FMR_lfw_facenet} and \ref{fig:01FMR_lfw_arcface} show the same on LFW.} \vspace{-2mm}
\label{fig:01FMR}
\end{figure*}

\begin{figure*}[h]
\centering
  \subfloat[Adience - FaceNet \label{fig:eer_adience_facenet}]{%
       \includegraphics[width=0.43\textwidth]{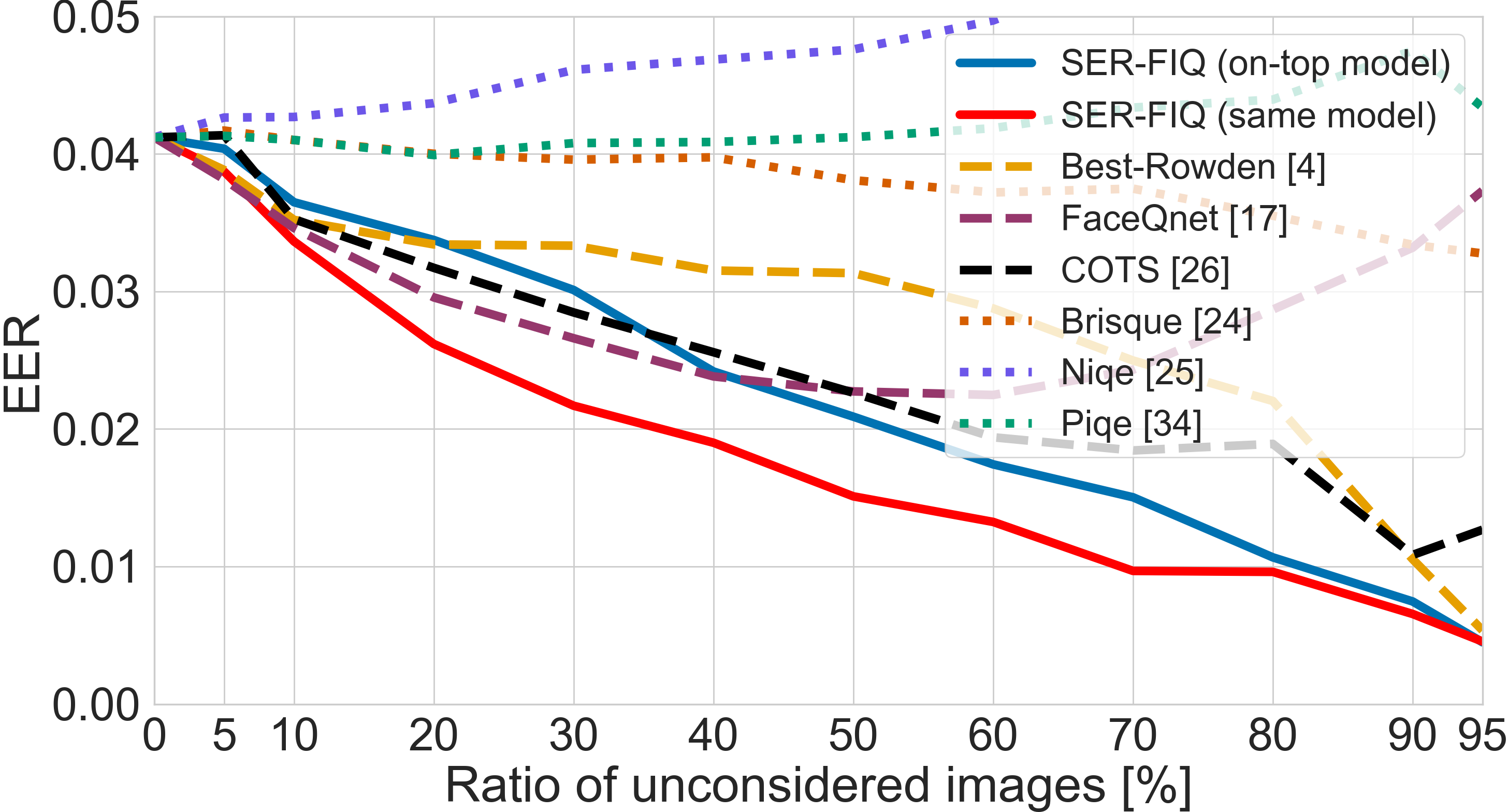}} \hspace{1mm}
  \subfloat[Adience - ArcFace \label{fig:eer_adience_arcface}]{%
       \includegraphics[width=0.43\textwidth]{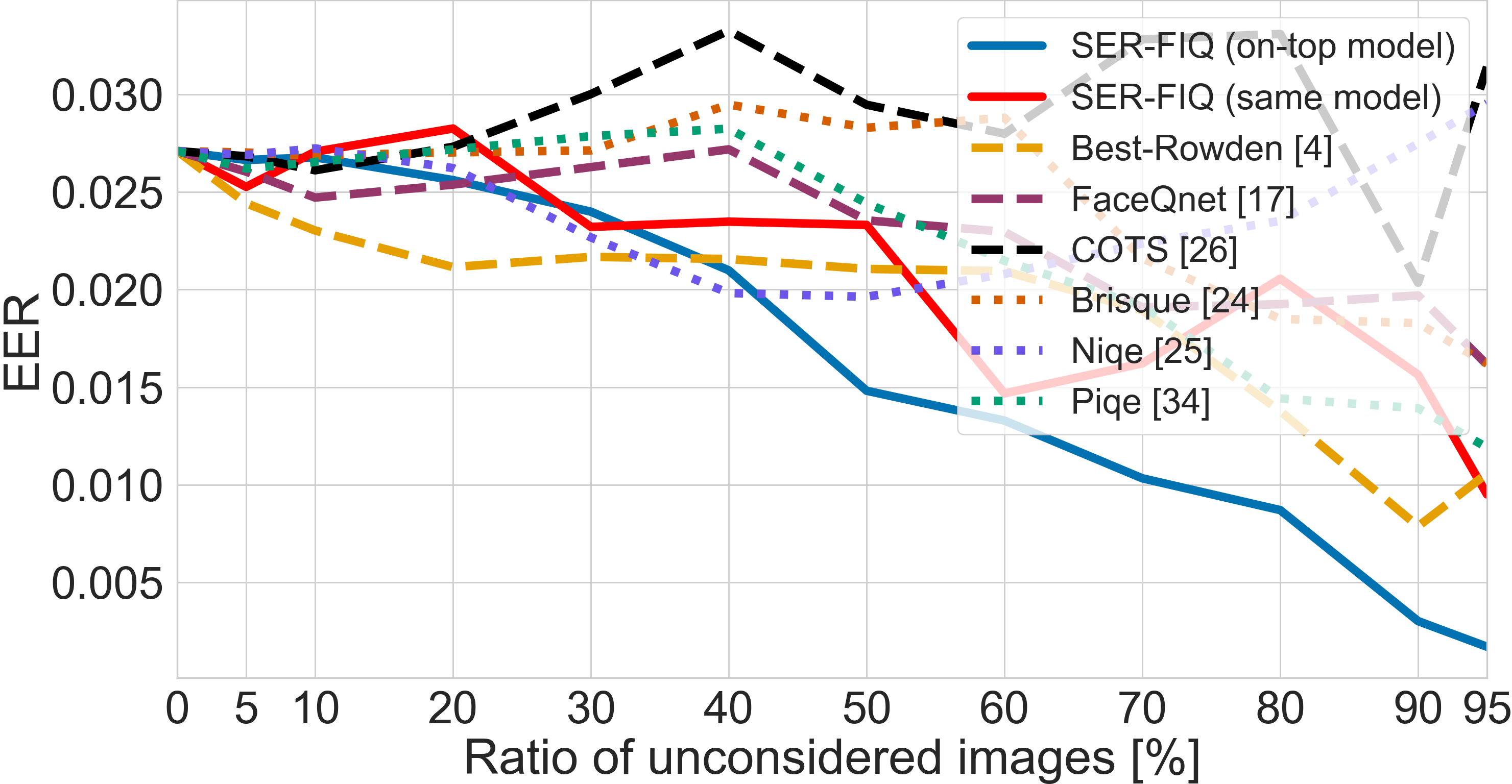}} \hspace{1mm}
  \subfloat[LFW - FaceNet \label{fig:eer_lfw_facenet}]{%
       \includegraphics[width=0.43\textwidth]{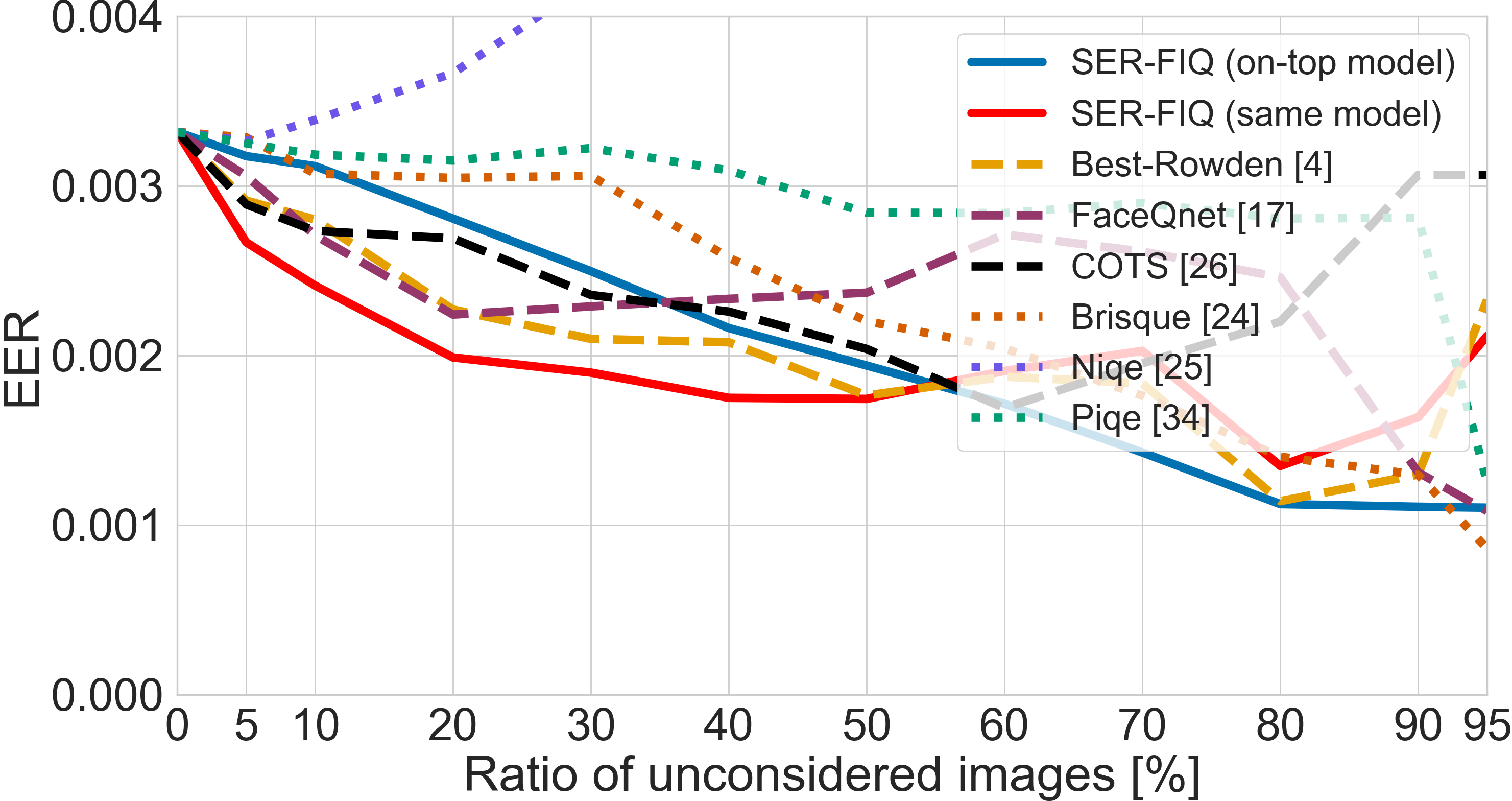}} \hspace{1mm}
  \subfloat[LFW - ArcFace \label{fig:eer_lfw_arcface}]{%
       \includegraphics[width=0.43\textwidth]{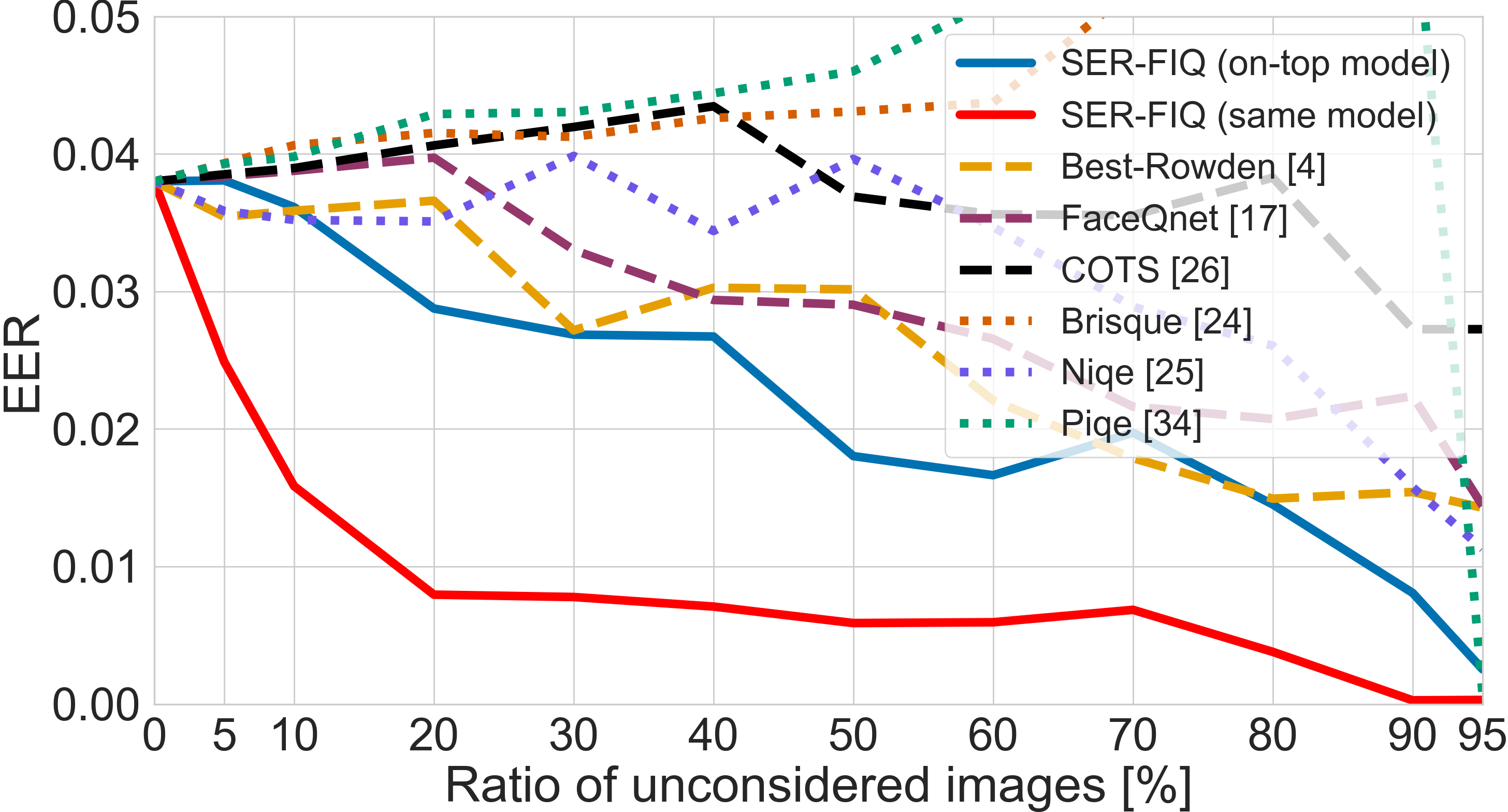}} \hspace{1mm}
\caption{The face verification performance given as EER for the predicted face quality values. The curves show the effectiveness of rejecting low-quality face images in terms of EER. Figure \ref{fig:eer_adience_facenet} and \ref{fig:eer_adience_arcface} show the results for FaceNet and ArcFace embeddings on Adience. Figure \ref{fig:eer_lfw_facenet} and \ref{fig:eer_lfw_arcface} show the some on LFW.} \vspace{-2mm}
\label{fig:EER}
\end{figure*}

The experiments are evaluated at three different operation points to investigate the face quality assessment performance over a wider spectrum of potential applications.
Following the best practice guidelines for automated border control of the European Border and Coast Guard Agency Frontex \cite{FrontexBestPractice}, Figure \ref{fig:001FMR} shows the face quality assessment performance at a FMR of 0.001.
Figure \ref{fig:01FMR} presents the same at a FMR of 0.01 and Figure \ref{fig:EER} shows the face quality assessment performance at the widely-used EER.
Moreover, Figure \ref{fig:SampleImages} shows sample images with their corresponding quality predictions.
Since the statements about each tested face quality assessment approach are very similar over all experiments, we will make a discussion over each approach separately.

\vspace{-0mm}

\paragraph{No-reference image quality approaches}
To understand the importance of different image quality measures for the task of face quality assessment, we evaluated three no-reference quality metrics  Brisque \cite{6272356}, Niqe \cite{6353522}, Piqe \cite{7084843} (all represented as dotted lines).
While in some evaluation scenarios the verification error decrease when the proportion of neglected images (low quality) is increased, in most cases they lead to an increased verification error.
This demonstrates that image quality alone is not suitable for generalized face quality estimation.
Factors of the face (such as pose, occlusions, and expressions) and model biases are not covered by these algorithms and might play an important role for face quality assessment.

\vspace{-2mm}
\paragraph{Best-Rowden}
The proposed approach from Best-Rowden and Jain \cite{DBLP:journals/corr/Best-RowdenJ17} works well in most scenarios and reaches a top-rank performance in some minor cases (e.g. LFW with FaceNet features).
However, it shows instabilities that can lead to highly wrong quality predictions.
This can be observed well on the Adience dataset using FaceNet embeddings, see Figure \ref{fig:001FMR_adience_facenet} and \ref{fig:01FMR_adience_facenet}.
These mispredictions might be explained by the ColorFeret training data that does not contain all important quality factors for a given face embedding.
On the other hand, these quality factors are generally unknown and thus, training data should never be considered to be covering all factors.

\paragraph{FaceQnet}
FaceQnet \cite{DBLP:journals/corr/abs-1904-01740}, proposed by Hernandez-Ortega et al., shows a suitable face quality assessment behaviour in most cases.
In comparison with other face quality assessment approaches, it only shows a mediocre performance.
Although FaceQnet was trained on labels coming from the same FaceNet embeddings as in our evaluation setting, it often fails in predicting well-suited quality labels on these embeddings, e.g. in Figure \ref{fig:001FMR_lfw_facenet} on LFW.
Also on Adience (e.g. Figure \ref{fig:01FMR_adience_facenet} and \ref{fig:eer_adience_facenet}), the performance plot shows a U-shape that demonstrates that the algorithm can not distinguish well between medium and higher quality face images.
Since the method is trained on the same features, these FaceNet-related instabilities might result from overfitting.

\paragraph{COTS}
The industry baseline COTS \cite{COTS} from Neurotechnology generally shows a good face quality assessment when the used face recognition system is based on FaceNet features.
Specifically on LFW (see Figure \ref{fig:001FMR_lfw_facenet}, \ref{fig:01FMR_lfw_facenet}, and \ref{fig:eer_lfw_facenet}) a small U-shape can be observed similar to FaceQnet.
While it shows a good performance using FaceNet embeddings, the face quality predictions using the more recent ArcFace embeddings are of no significance (see Figure \ref{fig:001FMR_adience_arcface}, \ref{fig:001FMR_lfw_arcface}, \ref{fig:01FMR_adience_arcface}, \ref{fig:01FMR_lfw_arcface}, \ref{fig:eer_adience_arcface}, and \ref{fig:eer_lfw_arcface}).
Here, rejecting face images with low predicted face quality does not improve the face recognition performance.
Since no information about the inner workflow is given, it can be assumed that their method is optimized to more traditional face embeddings, such as FaceNet.
More recent embeddings, such as ArcFace, are probably intrinsically robust to the quality factors that COTS is trained on.

\paragraph{SER-FIQ (on-top model)}
On the contrary to the discussed supervised methods, our proposed unsupervised solution that builds on training a small custom face recognition network shows a stable performance in all investigated scenarios (Figure \ref{fig:001FMR}, \ref{fig:01FMR}, and \ref{fig:EER}).
Furthermore, our solution is always close to the top performance and outperforms all baseline approaches in the majority of the scenarios, e.g. in Figure \ref{fig:001FMR_adience_facenet}, \ref{fig:001FMR_lfw_arcface}, \ref{fig:01FMR_adience_facenet}, \ref{fig:01FMR_adience_arcface}, \ref{fig:01FMR_lfw_arcface}, \ref{fig:eer_adience_facenet}, \ref{fig:eer_adience_arcface}, and \ref{fig:eer_lfw_arcface}.
Our method proved to be particularly effective in combination with recent ArcFace embeddings (see Figures \ref{fig:01FMR_adience_arcface}, \ref{fig:01FMR_lfw_arcface}, \ref{fig:eer_adience_arcface}, and \ref{fig:eer_lfw_arcface}).
The unsupervised nature of our solution seems to be a more accurate and more stable strategy. 


\paragraph{SER-FIQ (same model)}
Our method that avoids training by utilizing the deployed face recognition systems is build on the hypotheses that face quality assessment should aim at estimating the sample quality of a \textit{specific} face recognition model.
This way it adapts to the models' decision patterns and can predict the suitability of face sample more accurately.
The effect of this adaptation can be seen clearly in nearly all evaluated cases (see Figure \ref{fig:001FMR}, \ref{fig:01FMR}, and \ref{fig:EER}).
It outperforms all baseline approaches by a large margin and demonstrates an even stronger performance at small FMR (see Figures \ref{fig:001FMR_adience_facenet}, \ref{fig:001FMR_adience_arcface}, \ref{fig:001FMR_lfw_facenet}, and \ref{fig:001FMR_lfw_arcface} at the Frontex recommended FMR of 0.001). 
This demonstrates the benefit of focusing on the face quality assessment to a specific (the deployed) face recognition model.

\section{Conclusion}
Face quality assessment aims at predicting the suitability of face images for face recognition systems.
Previous works provided supervised models for this task based on inaccurate quality labels with only limited consideration of the decision patterns of the deployed face recognition system.
In this work, we solved these two gaps by proposing a novel unsupervised face quality assessment methodology that is based on a face recognition model trained with dropout.
Measuring the embeddings variations generated from random subnetworks of the face recognition model, the representation robustness of a sample and thus, the sample's quality is determined.
To evaluate a generalized face quality assessment performance, the experiments were conducted using three publicly available databases in a cross-database evaluation setting.
We compared our solution on two different face embeddings against six state-of-the-art approaches from academia and industry.
The results showed that our proposed approach outperformed all other approaches in the majority of the investigated scenarios.
It was the only solution that showed a consistently stable performance.
By using the deployed face recognition model for verification and the proposed quality assessment methodology, we avoided the training phase completely and further outperformed all baseline approaches by a large margin.


\paragraph{Acknowledgement} 
This research work has been funded by the German Federal Ministry of Education and Research and the Hessen State Ministry for Higher Education, Research and the Arts within their joint support of the National Research Center for Applied Cybersecurity ATHENE.
Portions of the research in this paper use the FERET database of facial images collected under the FERET program, sponsored by the DOD Counterdrug Technology Development Program Office.

\clearpage
\newpage

{\small
\bibliographystyle{ieee_fullname}
\bibliography{egbib}
}

\end{document}